\documentclass{article}

\usepackage{PRIMEarxiv}

\usepackage[utf8]{inputenc} 
\usepackage[T1]{fontenc}    
\usepackage[pagebackref,breaklinks,colorlinks]{hyperref}
\usepackage{url}            
\usepackage{booktabs}       
\usepackage{amsfonts}       
\usepackage{nicefrac}       
\usepackage{microtype}      
\usepackage{fancyhdr}       
\usepackage{graphicx}       

\usepackage{amsmath}
\usepackage{algorithmic}
\usepackage{algorithm}
\usepackage{array}
\usepackage[caption=false,font=normalsize,labelfont=sf,textfont=sf,position=top]{subfig}
\usepackage{bm}
\usepackage{textcomp}
\usepackage{stfloats}
\usepackage{verbatim}

\usepackage{multirow}
\usepackage{subcaption}
\usepackage{threeparttable}
\usepackage{color}
\usepackage{xcolor}
\usepackage{colortbl}
\usepackage{svg}
\usepackage{pifont}
\usepackage{tikz}
\usepackage{siunitx}

\usetikzlibrary{shapes.geometric,shapes.symbols}

\graphicspath{{media/}}     

\newcommand{\tikzsymbol}[3][circle]{\tikz[baseline=-0.5ex]\node[inner
	sep=#3,shape=#1,draw,#2]{};}%

\newcommand{\cmark}{\ding{51}}
\newcommand{\tablestyle}[2]{\setlength{\tabcolsep}{#1}\renewcommand{\arraystretch}{#2}\centering\footnotesize}

\definecolor{tablered}{RGB}{205,51,51}
\definecolor{tablegreen}{HTML}{39b54a}
\definecolor{tableblue}{HTML}{4682B4}
\definecolor{tablegrey}{HTML}{808080}

\definecolor{figgrey}{RGB}{128,128,128}
\definecolor{figcyan}{RGB}{0,139,139}
\definecolor{figdarkred}{RGB}{64,0,0}
\definecolor{figred}{RGB}{205,51,51}
\definecolor{figgreen}{RGB}{113,198,113}

\definecolor{tomato}{HTML}{FF6347}
\definecolor{royalblue}{HTML}{4169E1}
\definecolor{springgreen}{HTML}{00FF7F}

\definecolor{simcolor1}{RGB}{214,96,77}
\definecolor{simcolor2}{RGB}{67,147,195}
\definecolor{rectcolor}{RGB}{27,120,55}

\title{Few-Shot Video Object Segmentation in X-Ray Angiography Using Local Matching and Spatio-Temporal Consistency Loss
\thanks{\textit{\underline{Citation}}:
\textbf{Xi, L., Ma, Y., and Zhuang. X., 2026. Few-shot video object segmentation in X-ray angiography using local matching and spatio-temporal consistency loss. \textit{Neural Networks}, 200, p.108808.}} 
}

\author{
  Lin Xi$^{1}$,~~Yingliang Ma$^{1,2,}$\thanks{Corresponding author},~~Xiahai Zhuang$^{3}$\\
  $^{1}$University of East Anglia,~~United Kingdom\\
  $^{2}$King's College London,~~United Kingdom\\
  $^{3}$Fudan University,~~China\\
  \texttt{xilin.chibchin@outlook.com},~~\texttt{yingliang.ma@uea.ac.uk}\\
}

\begin{document}
\maketitle

\begin{abstract}
	High-quality, densely annotated data serve as a crucial foundation for developing robust X-ray angiography segmentation models. However, obtaining per-object pixel-level annotations in the medical domain is both expensive and time-consuming, often requiring close collaboration between clinical experts and developers. This paper aims to reduce the annotation costs of X-ray angiography videos by leveraging few-shot video object segmentation (FSVOS), which separates target objects from the background using only a single annotated frame during inference. We introduce a novel FSVOS model that employs a local matching strategy to restrict the search space to the most relevant neighboring pixels. Rather than relying on inefficient standard im2col-like implementations (\emph{e.g.}, spatial convolutions, depthwise convolutions and feature-shifting mechanisms) or hardware-specific CUDA kernels (\emph{e.g.}, deformable and neighborhood attention), which often suffer from limited portability across non-CUDA devices, we reorganize the local sampling process through a direction-based sampling perspective. Specifically, we implement a non-parametric sampling mechanism that enables dynamically varying sampling regions. This approach provides the flexibility to adapt to diverse spatial structures without the computational costs of parametric layers and the need for model retraining. To further enhance feature coherence across frames, we design a supervised spatio-temporal contrastive learning scheme that enforces consistency in feature representations. In addition, we introduce a publicly available benchmark dataset for multi-object segmentation in X-ray angiography videos (MOSXAV), featuring detailed, manually labeled segmentation ground truth. Extensive experiments on the CADICA, XACV, and MOSXAV datasets show that our proposed FSVOS method outperforms current state-of-the-art video segmentation methods in terms of segmentation accuracy and generalization capability (\emph{i.e.}, seen and unseen categories). This work offers enhanced flexibility and potential for a wide range of clinical applications. Code is available at: \href{https://github.com/xilin-x/XRAVOS}{https://github.com/xilin-x/XRAVOS}
\end{abstract}


\section{Introduction}\label{sec:intro}

X-ray angiography video segmentation of blood vessels and other surgical objects is a critical step in 3D vessel reconstruction, coronary artery analysis, and cardiac modeling. Frame-wise and pixel-wise segmentation enables precise localization of each object in X-ray videos, aiding in surgical planning, biomarker computation, and various downstream tasks. However, in many practical scenarios, obtaining dense annotations from clinical experts, especially pixel-level annotations for each frame, is labor-intensive and time-consuming. Moreover, the limited availability of samples for rare anomalies and unusual pathological conditions further complicates the annotation process.

\begin{figure}[!t]
	\begin{center}
		\includegraphics[width=0.6\columnwidth]{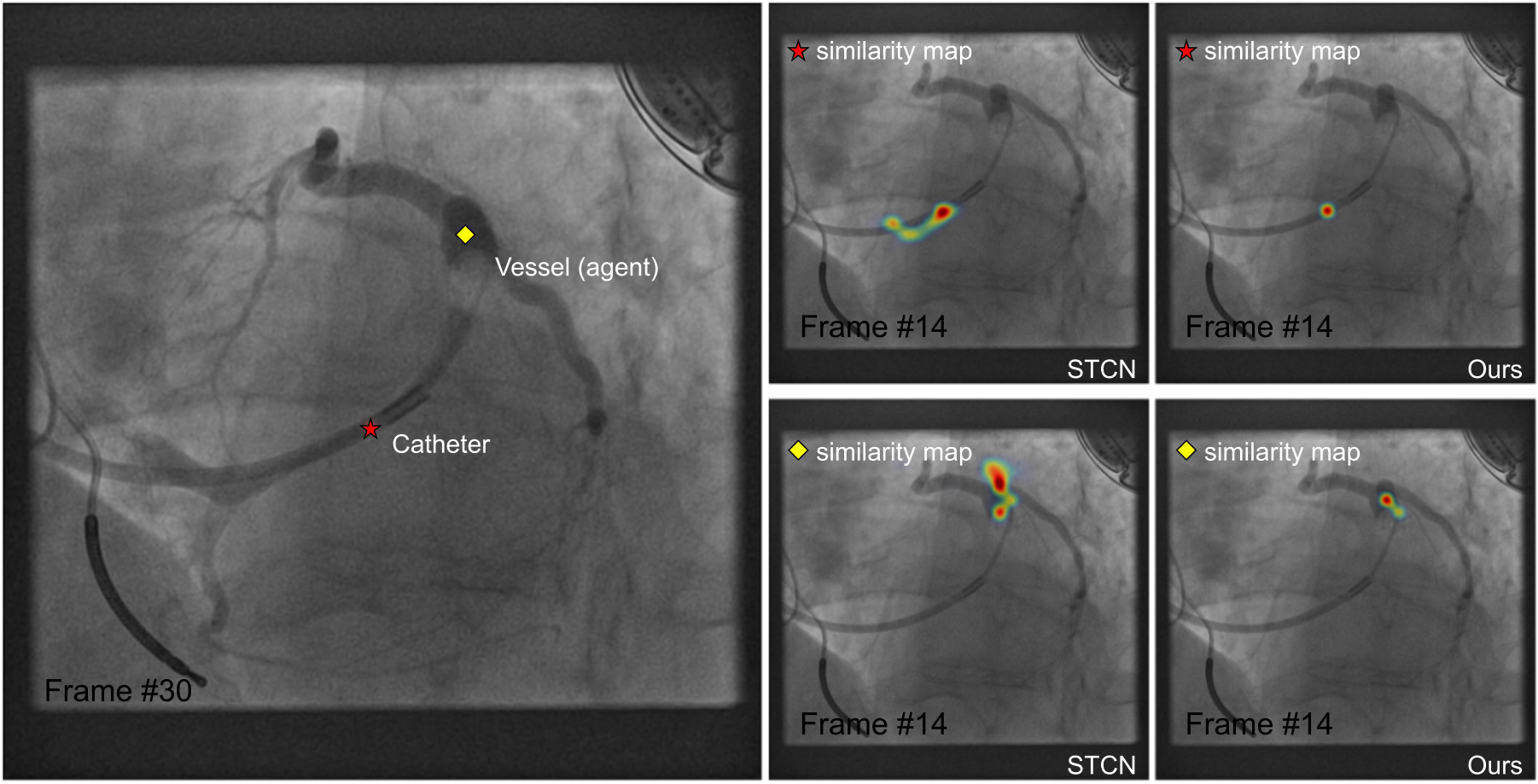}
	\end{center}
	\caption{Visualization of similarity maps between the query frame (\#30) and the support frame (\#14). Reference labels manually selected in the query frame (leftmost) are used to compute similarity scores relative to the support frame. In the resulting heatmaps, regions in deep red indicate higher similarity. The yellow diamond (\tikzsymbol[diamond]{fill=yellow}{1.5pt}) denotes the coronary vessels, while the red star (\tikzsymbol[star]{fill=red,star point ratio=2.618}{1pt}) identifies the injection catheter.}
	\label{fig:head}
\end{figure}

Driven by the paradigm of class-agnostic mask tracking, the computer vision community has increasingly focused on few-shot video object segmentation (FSVOS) \cite{OSVOS,MaskTrack,STM,STCN,XMem,RAB,CENet}. This shift prioritizes robust tracking mechanisms over the explicit learning of object-specific semantic representations, thereby mitigating the heavy reliance on large-scale annotated datasets. The goal of FSVOS is to segment target objects throughout an entire video sequence based on an initial object mask, which can be either manually provided or automatically generated from any single frame. As a widely accepted solution for FSVOS, matching-based methods \cite{STM,STCN,XMem} perform a matching strategy to generate dense correspondences between the query frame and the support frame(s), where the past frames as well as corresponding masks are stored in an external memory to build a feature memory \cite{AOT}. In Space-Time Memory (STM) network \cite{STM}, cross-image correspondences is the key component that makes it superior in performance and simplicity. In the pixel-wise retrieval phase, it enables the global content-dependent interactions among different image regions to model long-range dependencies. Through the visualization of dense space-time correspondence, we observe the crucial role of local contexts, such as the neighborhood pixels \cite{SASA,Swin,Focal,NAT,SlideTransformer}, in establishing correspondences for cross-frame matching (see the similarity scores in Fig. \ref{fig:head}). Meanwhile, compared to local-side computing solutions \cite{TII2025Zhang}, global and fine-grained dependencies become less critical due to spatial redundancy and noise as the number of features involved in non-local matching increases, particularly in cases of locally recurring motion such as cardiac and respiratory movements, or within specific regions of interest in X-ray angiography videos. On the other hand, matching-based methods \cite{STM,STCN,XMem,LiVOS} focus only on discovering space-time correspondences from inter-frames, while ignoring the valuable temporal consistency of object representation across multiple frames. Indeed, due to only mining the matched pixels across the memory frames, some background pixels/patches are wrongly recognized as highly correlated to the query primary objects. Therefore, it is necessary to make full use of inter-frame consistency to make up for the drawbacks caused by ignoring temporal coherence.

\begin{figure}[!t]
	\begin{center}
		\includegraphics[width=0.5\columnwidth]{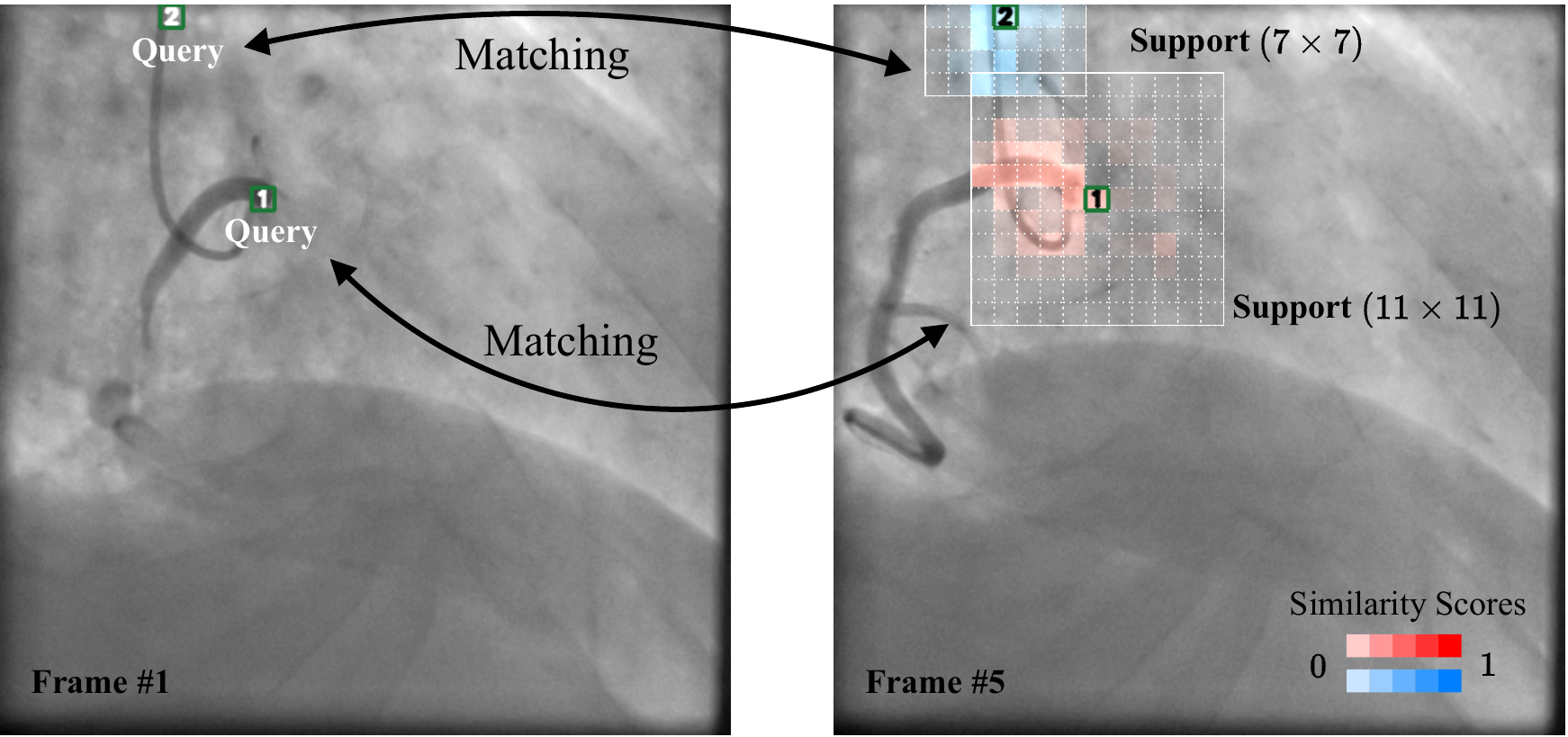}
	\end{center}
	\caption{An illustration of our local matching between the query frame and the support frame at the given patch (position \textbf{1} and \textbf{2}). Position \fcolorbox{rectcolor}{tablegrey}{\textbf{1}} and \fcolorbox{rectcolor}{tablegrey}{\textbf{2}} on the key frames correspond to the spatial positions of \fcolorbox{rectcolor}{tablegrey}{\textcolor{white}{\textbf{1}}} and \fcolorbox{rectcolor}{tablegrey}{\textcolor{white}{\textbf{2}}} in the current frame. The similarity scores are represented by colorful heatmap masks (\emph{i.e.}, \tikzsymbol[rectangle]{fill=simcolor1,minimum width=4pt}{2pt} and \tikzsymbol[rectangle]{fill=simcolor2,minimum width=4pt}{2pt}), where darker colors indicate higher similarity.}
	\label{fig:op_sample}
\end{figure}

In the domain of vision Transformers, window-based self-attention \cite{Swin,Focal} has emerged as a powerful and efficient mechanism for capturing local dependencies. The Swin Transformer \cite{Swin} partitions feature maps into non-overlapping windows and applies self-attention to each independently, leveraging batched matrix multiplication for high parallelization. Similarly, the Focal Transformer \cite{Focal} employs fine-grained local attention to account for short-range visual dependencies. These architectural advancements provide a strong motivation to model localized visual dependencies as a means of mitigating the high computational costs inherent in existing global matching-based methods \cite{STM,STCN,XMem,LiVOS}. Current paradigms based on the matching framework generate global correlations to facilitate segmentation. While effective, a significant limitation of these methods is that fine-level local interactions between the query and the memory become increasingly diluted as the memory bank grows. This accumulation of memory pixels introduces noise and reduces the signal-to-noise ratio, ultimately limiting the model's capacity to extract highly accurate and well-localized correspondences. A parallel challenge exists in standard Transformer architectures, where global self- and cross-attention mechanisms often struggle to preserve fine-grained spatial details when processing large feature maps. Inspired by local attention mechanisms \cite{SASA,NAT,SlideTransformer}, we explore local correspondences within a relevant neighborhood of sampling locations to enhance the robustness of feature matching. Furthermore, to improve feature coherence, we reinforce spatio-temporal correspondences by enforcing globally consistent feature representations across frames, ensuring that local precision does not come at the expense of global context.

In this work, we propose a novel FSVOS framework for X-ray angiography videos that captures local correspondences during the feature retrieval process between query and support frames. Since visual dependencies are typically stronger among nearby regions than distant ones, we restrict the correspondence search to a neighborhood of sampling locations in the support frames, as depicted in Fig. \ref{fig:op_sample}. By focusing on relevant local regions rather than the global field, this design improves the robustness of fine-grained matching while maintaining a low computational cost. Each query feature attends to its closest neighborhood in the support frames at the same granularity, effectively covering the most relevant regions while substantially reducing the number of spatio-temporal computations compared to dense matching mechanisms such as \cite{STCN,XMem}. Moreover, as shown in Fig.~\ref{fig:head}, global matching strategies often attend to uninformative background regions, causing salient foreground objects to be submerged in interference signals. From an implementation prespective, existing local-attention and matching approaches suffer from notable practical limitations. On one hand, methods based on standard im2col-like implementations \cite{SASA,SlideTransformer,Swin,Focal} exhibit sub-optimal memory efficiency and inefficient data access patterns. Specifically, Stand-Alone Self-Attention (SASA) \cite{SASA} and Slide Attention \cite{SlideTransformer} are typically implemented using generic spatial or depth-wise convolutions. These designs incur redundant costs in kernel parameter storage and loading, leading to inefficient memory utilization. Similarly, Swin Transformer \cite{Swin} and Focal Transformer \cite{Focal} adopt window-based partitioning with cyclic shifts executed through reshape and roll operations. While effective for batch training, these shifts introduce non-adjacent spatial regions into the same attention window, requiring specialized attention masks to preserve local receptive fields. As a result, substantial intermediate data movement and increased memory consumption are often unavoidable. In contrast, our framework leverages a non-parametric sampling operation that avoids these architectural inefficiencies and hardware dependencies. This design ensures no additional inference-time overhead compared to existing matching-based models, facilitating seamless deployment across diverse clinical computing environments, including the CPU-only systems commonly found in hospital settings. To further enforce spatio-temporal consistency, we incorporate a supervised contrastive learning scheme during traning. By defining object centers based on ground-truth labels, we encourage pixels belonging to the same object to form compact feature clusters while maintaining clear separation from other objects. This framework explicitly links query-frame features with cross-frame global representations. Crucially, this scheme requires no additional pixel-wise annotations and incurs no extra cost during inference, making the overall framework both data-efficient and computationally practical.

We validate the effectiveness of our method by evaluating it on various public X-ray angiography benchmark video datasets, \emph{i.e.}, CADICA \cite{CADICA}, XACV \cite{XACV}, and MOSXAV. On all datasets, our method achieves a competitive performance compared to the current state-of-the-art methods while maintaining a real-time inference speed. In particular, instead of a global matching mechanism, our approach uses local matching to effectively explore pixel dependencies compared with previous matching-based solutions \cite{STM,STCN,XMem} in the X-ray angiography video segmentation. Note that our all our experiments are implemented on a single NVIDIA RTX\texttrademark\ 6000 GPU and CPU platform, which makes our method much more accessible than other state-of-the-art methods, which require powerful hardware resources. Our main contributions can be summarized as follows:
\begin{itemize}
	\item We propose a novel FSVOS method for X-ray angiography video by exploiting the most relevant neighborhood region.
	\item To make the sampling process simpler and more flexible, we propose a non-parametric sampling method that enables dynamically adjustable sampling regions without requiring retraining.		
	\item We introduce a new benchmark dataset\footnote{https://github.com/xilin-x/MOSXAV}, the X-ray Multi-Object Segmentation in X-ray Angiography Videos (MOSXAV). MOSXAV focuses on understanding multiple objects in X-ray videos, offering high-quality, manually labeled segmentation ground truth.
\end{itemize}

\section{Related works}\label{sec:rw}

\subsection{Vessel and catheter segmentation}

Vessel segmentation is an active research topic in medical image analysis. Among the most widely used backbones, U-Net \cite{UNet} has recently become the most popular. To address the limitation of U-Net, which only allows a single set of concatenation layers between encoder and decoder blocks, T-Net \cite{T-Net} introduces a nested encoder-decoder architecture. This design employs multiple small encoder-decoder sub-networks for each block of the convolutional network, enabling more effective vessel segmentation in coronary angiography. Sequential-image-based approaches, such as SVS-Net \cite{SVSNet}, adopt an encoder-decoder architecture that leverages multiple contextual frames of 2D images to improve segmentation of 2D vessel masks. Khan \emph{et al.} \cite{KHAN} proposed a contextual network for accurate retinal vessel identification, with particular consideration of computational efficiency to support deployment on resource-constrained devices such as smartphones. Recent methods have also integrated ViT \cite{ViT} with graph neural networks or convolutional block attention modules to improve structural understanding of vascular morphology, as exemplified by G2ViT \cite{G2ViT} and CBAM\&ViT \cite{CBAMViT}.

In addition, automatic catheter segmentation has emerged as another active topic in medical image analysis. Existing work has reported catheter segmentation in X-ray images, including electrophysiology (EP) electrodes \cite{Baur2016MIAR} and EP catheters \cite{Ma2015TMI,Ambrosini2017MICCAI,CathMTL}. More recently, Kim \emph{et al.} \cite{KIM2022CMIG} addressed the challenge of detecting active acoustic catheters (AACs) in echocardiography. Their method first applies U-Net to segment the left ventricle, then filters the color response generated by the AAC, and finally performs thresholding to identify the catheter. Yang \emph{et al.} \cite{yang2019catheter} proposed a 3D ultrasound catheter segmentation method based on Shared-ConvNet, which employs CNNs with shared weights across imaging planes and performs voxel-wise binary classification. Overall, beyond our prior work, the literature on catheter segmentation in AI-driven, ultrasound-guided minimally invasive endovascular surgery (MIES) remains limited. Nguyen \emph{et al.} \cite{Nguyen2020ICRA} introduced an end-to-end, real-time deep learning framework for endovascular intervention, incorporating a novel flow-guided warping function to enforce frame-to-frame temporal continuity. To capture such sequential behavior, Ranne \emph{et al.} \cite{Ranne2024ICRA} proposed the Attention-in-Attention ResNet for Segmentation (AiAReSeg), which aggregates information across image sequences to infer knowledge about the current frame.

\subsection{Few-shot video object segmentation}

FSVOS \cite{OSVOS} differs from fully automated video segmentation \cite{COSNet,IMCNet,ClusterNet} by incorporating human input, specifically a first-frame segmentation mask of the target objects. The methods then segment the desired objects in the remaining frames. Here, few-shot refers to the level of human interaction at test time, not during the training phase. The few-shot methods still rely on the supervised learning paradigm to train a pixel-wise tracking \cite{STM,STCN,XMem,LiVOS} or mask propagation framework. Early few-shot methods \cite{OSVOS,OSVOSS} employ online fine-tuning on the basis of this supervision, which suffered from high test runtime and were gradually phased out. The follow-up researches have been explored including embedding learning \cite{CFBI,FEELVOS}, propagation-based \cite{MaskTrack,RVOS} and tracking-based \cite{SiamMask,MASKRNN}. These methods perform frame-to-frame propagation or global matching with the first reference frame, while the context is still limited and it becomes harder to match as the video progresses. To fully exploit the spatial-temporal context over longer distances, recent state-of-the-art methods define the past frames with object masks as feature memory and the current frame as the query \cite{STM,FRTM}. As a representative work, STM \cite{STM} utilizes an external memory to store the past frames as well as corresponding masks, where a dense matching is adopted to establish long-range dependencies in order to achieve pixel classification and objects segmentation. Staring from STM \cite{STM}, matching-based paradigm was adopted and has been extended by many follow-up works \cite{MiVOS,STCN,XMem,LiVOS}, which achieve leading accuracy on most benchmarks duo to long-range context support.

The latest research in matching-based FSVOS has primarily concentrated on enhancing network architectures, such as STCN \cite{STCN}, XMem \cite{XMem}, and LiVOS \cite{LiVOS}. STCN \cite{STCN} employs an effective and efficient matching strategy for establishing dense correspondences between query and support frame(s), eliminating the need for re-encoding the mask features for every objects, as observed in STM \cite{STM}. These methods often prioritize the exploration of valid global correspondences while disregarding the shaping of the discriminative embedding space. Moreover, our contribution builds upon these studies, as we have not only developed a more efficient matching model based on the current matching-based FSVOS methods \cite{STCN,XMem,LiVOS} but also made advancements in ensuring consistency between query and memory in the aspect of feature learning.

In medical data analysis, there has been increasing interest in FSVOS techniques that enable class-agnostic mask tracking and segmentation using only a few annotated examples. For instance, RAB \cite{RAB} is a two-phase framework that introduces a spatio-temporal consistency relearning strategy to adapt an image segmentation model for video data. In contrast, our approach is a one-stage, matching-based method that not only develops a more efficient matching strategy but also strengthens the consistency between query and support frames through improved feature learning.

\subsection{Transformer architectures}

Transformer architectures are a type of neural network architecture that have proven extremely adept at modelling long-term relationships within an input sequence via self-attention mechanisms \cite{Transformer}. The exploration of global matching between query and memory in STM network and their variants closely resembles the self/cross-attention mechanisms found in Transformer. This exploration aims to uncover global correlations. Balancing interaction granularity is a persistent challenge for Transformer-based methods. Because of the significant computational burden associated with global interaction, the input features are often downsampled to a lower resolution, which to some extent restricts the networks` capability for fine-grained feature learning. 

Recently, many efforts have been made to address this problem, which can be roughly divided into two categories in practice. The first category involves employing local attention modules to alleviate the issues mentioned above. The most direct approach is constraining the attention pattern to fixed local windows, which is commonly adopted by many works \cite{ViT,Swin,Focal,RegionViT,NAT}. While restricting the attention pattern to a local neighborhood can reduce complexity, it also sacrifices global information. The second category is to learn data-dependent sparse attention. Inspired by deformable convolution \cite{DCN}, Deformable DETR \cite{DeformableDETR} directs its attention to a small fixed set of sampling points, predicted from the the feature of query elements. PnP-DETR \cite{PnP-DETR} presents a similar idea, where it samples fine foreground features and pools background features into a reduce size. Sparse DETR \cite{SparseDETR} sparsifies encoder tokens by using a learnable decoder attention map predictor, which is different from Deformable DETR's key sparsification method.

In our method, a local matching module is proposed to enable effective message passing between the query and support frame(s) within a limited computational budget, focusing on fine-grained, locally relevant regions. Similar to local attention mechanisms in transformers \cite{SASA,NAT,SlideTransformer}, the query captures visual dependencies for specific locations within its local neighborhood; however, in our case, these dependencies are established with the corresponding regions of the support frame(s). To further improve flexibility and efficiency, we employ a non-parametric sampling strategy that allows dynamically varying sampling regions without requiring retraining or device-specific support.

\subsection{Supervised contrastive learning}

Supervised contrastive learning draws on existing literature in self-supervised representation learning, metric learning, and supervised learning. Typically, the state-of-the-art family of models for self-supervised representation learning utilizes the paradigm known as contrastive learning \cite{wu2018unsupervised,Henaff20,Hjelm19,CMC,Sermanet18,ChenK0H20,Tschannen2020On}. In these works, the losses are inspired by noise contrastive estimation \cite{GutmannH10,MnihK13} or N-pair losses \cite{Sohn16}. Recently, supervised contrastive loss \cite{SCL} has been introduced as a novel extension to the contrastive loss function. This innovative approach allows for multiple positives per anchor, thus adapting contrastive learning to the fully supervised setting. Following \cite{SCL}, supervised contrastive loss has also been employed in various downstream tasks \cite{ZhangTRR21,ZhaoSS2021}. To enhance consistency between query and memory, we proposed a supervised contrastive learning scheme. In contrast to the traditional supervised contrastive loss \cite{SCL}, which does not explicitly enforce intra-class compactness, our approach defines object centers based on ground-truth labels, encouraging pixels of the same object to form compact feature clusters while preserving clear separation across different objects.

\section{Method}\label{sec:met}

\subsection{Network formulation}

\begin{figure*}[!tb]
	\begin{center}
		\includegraphics[width=0.7\linewidth]{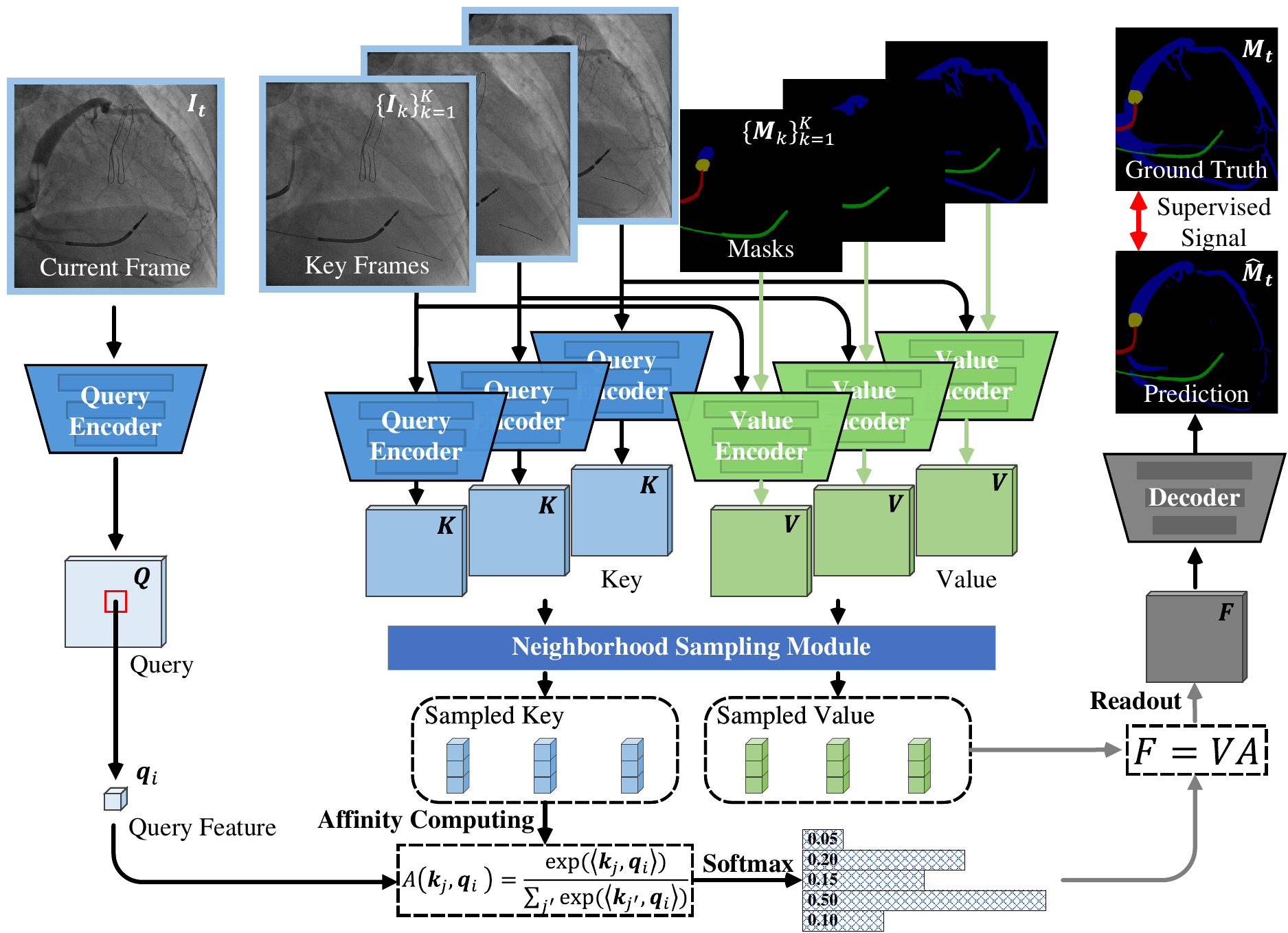}
	\end{center}
	\captionsetup{font=small}
	\caption{Overview of our proposed local matching-based FSVOS method. Given a query frame $\bm{I}_{t}$ and several key frames $\{\bm{I}_{k}\}_{k=1}^{K}$, sampled from a same training video, we first use local sampling function $\phi(\cdot;\cdot)$ to retrieve local sampled set $\Omega_{L}$ from key features $\bm{K}$ and value features $\bm{V}$. Then we make feature-level correspondence matching between query feature $\bm{Q}$ and sampled key features through Eq. \ref{eq:aff_com}-\ref{eq:sim_com}. Then we retrieve corresponding features $\bm{F}$ by aggregating sampled value features using the affinity matrix $\bm{A}$.}
	\label{fig:framework}
\end{figure*}

To modeling space-time correspondences in the context of X-ray video, our model architecture shares a similar multi-store feature memory design with \cite{STM,STCN}. The overall framework of the proposed method is shown in Fig. \ref{fig:framework}.

Given a video with $T$ frames $\mathcal{I}=\{\bm{I}_{t}\}_{1}^{T}$ and first-frame reference mask $\bm{M}_{1}$, our model match the object pixels and generates corresponding masks for subsequent query frames. For the current frame $\bm{I}_{t}$, we first fed it into the query encoder to extract the query feature $\bm{Q}\in\mathbb{R}^{HW\times D}$ in which the two spatial dimension were flattened. Subsequently, the spatial feature map $\bm{Q}$ is passed to the affinity module, where an affinity matrix $\bm{A}$ is obtained by applying a softmax function on a similarity matrix $\bm{S}$ between $\bm{Q}$ and the key feature $\bm{K}\in\mathbb{R}^{KNHW\times D}$, where $K$ is the number of the keyframes and $N$ is the number of local sampling elements. The similarity matrix $\bm{S}$ captures the pairwise similarities between each sampled key element (\emph{i.e.}, the most relevant neighborhood region) and every query element. Here, the local elements in the sampled key are generated through the proposed local sampling module, which samples key and value feature elements in parallel before storing them in memory. After obtaining the affinity $\bm{A}$, we retrieve the corresponding feature $\bm{F}\in\mathbb{R}^{HW\times C}$ from the external feature memory through memory reading, and then the readout feature $\bm{F}$ are employed in the generation of a segmentation mask for the current frame. In particular, the new memory elements which are generated by our proposed local sampling module can be added to the external memory when if the current frame is the keyframe. For the keyframe, we will take a fixed interval sample frame every $r$-th frame and identify it to keyframe in practice.

For the local sampling module, the inputs consist of the key and value features extracted from both the query and the value encoder. Within this module, fine-grained features are obtained from the corresponding neighborhood region of the keyframes' features according to specific locations of query element. The local sampling mechanism gathers tokens from both the key and value features in a consistent manner, thereby aggregating local contextual information. In contrast, standard STM network are designed to capture long-range dependencies at a fine-grained level, but they incur high computational costs when performing attention between the query and a large memory set. To address this limitation, we propose a local matching strategy that reduces the computational burden while preserving representational effectiveness.

\subsection{Local matching and aggregation}

The core idea of our matching and aggregation is to find a correlation matrix $\bm{C}$ between the current frame and the history to retrieve the corresponding features from the memory, which is more efficient and effective to establish dense correspondences for matching-based matching solution. Instead of attending to all key elements at a fine-grained level during matching, we proposed attending only to the most relevant neighborhood fine-grained key elements locally. As such, it can use as many sizes of the memory bank as the standard matching-based model but with much longer history coverage.

\subsubsection{Matching}\label{sssec:matching}

Consider a query $\bm{Q}\in\mathbb{R}^{HW\times D}$ and the keys $\bm{K}\in\mathbb{R}^{KHW\times D}$, an $i$-th query element is associated with a $j$-th key element in the memory, where $K$ is the number of the keyframes, \emph{i.e.}, $KHW=K\times H\times W$ --- we refer to them jointly $\mathcal{M}=\{(i, j),c_{ij}\}\subset\bm{Q}\times\bm{K}$ as the feature matches. The $i$ and $j$ index a query element $\bm{q}_{i}$ and a key element $\bm{k}_{j}$, as well as a similarity score $\bm{s}_{ij}$. 

As in the standard matching-based matching, the assignment $\mathcal{M}$ can be obtained by computing the correlation matrix $\bm{C}=\{c_{ij};i\in\Lambda, j\in\Omega\}$ for all possible matches, where $\Lambda$ and $\Omega$ specify the set of query and key elements, respectively. Meanwhile, instead of attending to all key/value elements in the key/value sampled set $\Omega$ for a specific query element $i$, we sample a total of $N$ elements from the keyframes (support) to construct the key $\bm{K}$ and the value $\bm{V}$, 
\begin{equation}\label{eq:key_set}
	\Omega=\Omega_{L},
\end{equation}
where $\Omega_{L}$ denotes the local sampled set, and $KN<KHW$.

In our case, the matching-based FSVOS is equivalent to solving a neighborhood voting problem, such as Nearest Neighbor Search (NNS) \cite{ANN}, which aggregate the matched representations by a similarity matrix $\bm{S}\in\mathbb{R}^{N\times HW}$ representing the weights.

\subsubsection{Affinity computing}

For clarity, we consider a single keyframe (\emph{i.e.}, $K$=1). The affinity matrix $\bm{A}$ is derived by applying a softmax operation along the memory dimension (rows) of the similarity matrix $\bm{S}$:
\begin{equation}\label{eq:aff_com}
	\bm{A}(\bm{k}_{j},\bm{q}_{i})=\mathrm{softmax}(\bm{S}(\bm{k}_{j},\bm{q}_{i})),
\end{equation}
where $i$ and $j$ denote the index of the query element and the key element. The similarity matrix is computed by
\begin{equation}\label{eq:sim_com}
	\bm{S}(\bm{k}_{j},\bm{q}_{i})=\left<\bm{k}_{j},\bm{q}_{i}\right>,
\end{equation}
where $\left<\cdot,\cdot\right>$ denotes a similarity measure, \emph{i.e.}, $\ell_{2}$ distance. Given the $i$-th query element $\bm{q}_{i}$ obtained from the query frame via the query encoder, the sampled key elements $\{\bm{k}_{j};j\in\Omega\}$ are collected from the local set, which is derive from the memory. The total number of key elements for the $i$-th query element is $N$.

\subsubsection{Local sampling}

\begin{figure}[!tb]
	\begin{center}
		\includegraphics[width=0.55\columnwidth]{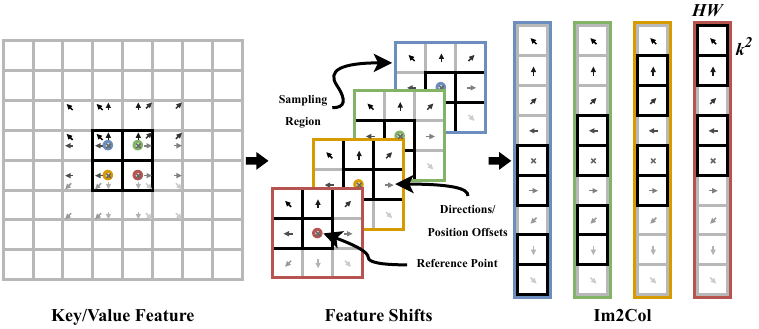}
	\end{center}
	\captionsetup{font=small}
	\caption{An illustration of our local sampling operation at feature level.}
	\label{fig:sampling}
\end{figure}

To efficiently search for the most relevant neighborhood region, we proposed a \textit{local sampling module}. Similar to the local attention mechanism in ViT's variants, we retrieve a small relevant subset $\Omega_{L}\subset\Omega$ from all spatial locations as the local set. Specifically, we define the sampling function $\phi(\cdot;\cdot)$ to attends only to a small set of sampling points around a reference point, independent of the spatial size of the feature maps, as shown in Fig. \ref{fig:sampling}. By assigning a fixed, limited number of keys to each query, this approach helps mitigate issues related to convergence and feature spatial resolution.

To achieve efficient affinity computation, the sampled local set needs to be packed into a matrix to enable parallel processing on accelerators. For that, the im2col operation \cite{HPCNN,Caffe} reorganizes image patches into columns of a matrix to implement general matrix multiply (GEMM), which can be efficiently executed using optimized basic linear algebra subprograms (BLAS) libraries. As depicted in Fig. \ref{fig:sampling}, the sampling region is centered at a specific query position (\emph{i.e.}, the reference point in Fig. \ref{fig:sampling}) and represents the region corresponding to its key/value pairs in memory. The sampled set is then flattened into columns, forming the final key/value matrix $\bm{M}$, which can be expressed as:
\begin{equation}\label{eq:im2col}
	\begin{aligned}
		\bm{M}[i,j]=\phi(\mathrm{p}(i), \Delta\mathrm{p}(j))=\bm{X}[\mathrm{p}(i)+\Delta\mathrm{p}(j)],\\
		\textrm{with}~i\in[0, HW-1]~\textrm{and}~j\in[0, k^{2}-1]~~~
	\end{aligned}
\end{equation}
where $\mathrm{p}(\cdot)$ represents the position function, and $\Delta\mathrm{p}(\cdot)$ is sampling direction (\emph{i.e.}, position offsets) determined by a regular grid $\mathcal{R}$ over the input feature map $\bm{X}$, where $\bm{X}$=$\{\bm{K}, \bm{V}\}$. The grid $\mathcal{R}$ defines the neighborhood region; for instance, when $k$=3, it is given by:
\begin{equation}\label{eq:neighborregion}
	\mathcal{R}=\{(-1, -1), (0, -1),\dots, (0, 1), (1, 1)\}
\end{equation}
which corresponds to a 3$\times$3 region with 9 possible directions. Notably, the sampling process primarily involves independently slicing the feature map according to the sampling windows from the sampling reference point perspective, as illustrated by the colored windows in Fig. \ref{fig:sampling}, similar to the approach used in SASA \cite{SASA} and NAT \cite{NAT}.

However, the above im2col operation can also be reinterpreted from a sampling direction-based perspective. In this view, the different directions, as illustrated in Fig. \ref{fig:sampling}, determine the structure of the key/value matrix, which contains $k^2$ rows. Each row corresponds to shifting the feature map in a specific direction, allowing us to reformulate the above equations as:
\begin{equation}\label{eq:dbasedview}
	\bm{M}[:,j]=\phi(:, \Delta\mathrm{p}(j)),\forall i,
\end{equation}
which is equivalent to shifting the original feature map in a specific direction defined by $\mathcal{R}$.

Instead of using depth-wise convolution with predefined fixed kernels as a substitute for feature shifts, we directly apply the non-parametric unfold operation, which enables dynamically varying sampling regions without requiring retraining. In the depth-wise convolution-based approach, the learnable fixed-shaped weights and biases are stored as model parameters during training. In contrast, our proposed unfold-based solution leverages a non-parametric sampling process, allowing it to seamlessly adapt to arbitrary sampling regions. Eq. \ref{eq:dbasedview} becomes
\begin{equation}\label{eq:unfold}
	\begin{aligned}
		\phi(i, \Delta\mathrm{p}(j))&=\mathrm{unfold}(:, \Delta\mathrm{p}(j))\\
		&=\bm{X}[\mathrm{p}(i)+\Delta\mathrm{p}(j)],\forall i,j.
	\end{aligned}
\end{equation}

In general, an efficient local sampling can be implemented from a direction-based perspective by carefully defining the sampling directions $\mathcal{R}$ within the unfold operation. This approach reduces the main computational overhead by avoiding inefficient slicing operations and leveraging optimized unfold operations on diverse hardware platforms \cite{HPCNN,cuDNN}.

Finally, to identify the most relevant neighborhood region for each query element, it is necessary to determine the appropriate sampling reference position(s) $\mathrm{p}(i)$ in Eq. \ref{eq:unfold}. In Focal Transformer \cite{Focal} and Slide Transformer \cite{SlideTransformer}, the query and key features are extracted from identical spatial locations within the same image space. Consequently, these methods directly use the spatial position of the query element as the sampling reference point, \emph{i.e.}, $\mathrm{p}(i)$$=$$i$, where $i$ denotes the spatial index of the query feature map. Under this formulation, each query element attends to its corresponding local neighborhood region in the keyframe feature map. However, considering that video frames often exhibit spatial dynamics, including both rigid and non-rigid motion, and that no explicit supervision is available during training, we further enhance the flexibility of the sampling process by incorporating the notion of cross-image similarity, as proposed in \cite{CCAM}. Specifically, the sampling reference point is determined based on appearance similarity between object representations across frames, enabling more adaptive and robust neighborhood aggregation. Given a query element $\bm{q}_{i}$, we compute its feature similarity with key elements $\{\bm{k}_{j};j\in\Omega\}$ using Eq. \ref{eq:sim_com} and identify the key element $\bm{k}_{j^{*}}$ that yields the highest similarity score:
\begin{equation}\label{eq:refpoint}
	j^{*}=\arg\max_{j\in\Omega}\bm{S}(\bm{k}_{j},\bm{q}_{i}).
\end{equation}
The sampling reference point $\mathrm{p}(i)$ is then defined as the spatial position of the key element $\bm{k}_{j^{*}}$ that yields the highest similarity score:
\begin{equation}\label{eq:refpoint_final}
	\mathrm{p}(i)=\mathrm{pos}(\bm{k}_{j^{*}}),
\end{equation}
where $\mathrm{pos}(\cdot)$ denotes the spatial position of the key element in the feature map, and the similarity scores are filtered using a top-$k$ operation. This design allows each query element to attend to the most relevant neighborhood region in the keyframe feature map based on appearance similarity, rather than solely relying on spatial proximity.

In summary, standard im2col-like implementations (\emph{e.g.}, spatial convolutions, depthwise convolutions and feature-shifting mechanisms) are data-rearrangement techniques that transform local neighborhoods into explicit, redundant patches to enable dense matrix multiplication. This process significantly increases memory consumption, as the footprint is proportional to the kernel or window size. In contrast, our direction-based unfolding treats local sampling as a logical offset mapping. By directly computing correlations between query features and sampled feature maps (or via index-based sampling), we eliminate the requirement for an intermediate ``flattened'' buffer. This results in a more memory-efficient and computationally streamlined implementation that preserves the inductive bias of local sampling without the overhead of physical data duplication.

\subsubsection{Aggregation}

Similar to \cite{STM}, we retrieve corresponding features $\bm{F}$ by aggregating sampled value features $\bm{V}$ using the affinity matrix $\bm{A}$ representing a readout operation that is controlled by the sampled key $\bm{K}$ and the query $\bm{Q}$,
\begin{equation}
	\bm{F}=\bm{V}\bm{A}(\bm{K},\bm{Q}).
\end{equation}
The retrieving operation maps every query element to a distribution over all $N$ memory elements and correspondingly aggregates their values $\bm{V}$. The feature $\bm{F}$ represents the information stored in the memory is aggregated via the matching correlation score $\bm{C}$. In the following, $\bm{F}$ is fed into the decoder to generate the mask, as illustrated in Fig. \ref{fig:framework}. At the next frame, the current frame with its predicted mask can be further added into the memory as new reference, and the query $\bm{Q}$ is reused as the key feature.

\subsection{Object-Aware Contrastive Learning}

In the general case, the existing matching-based approaches \cite{STM,STCN,XMem} mainly focus on feature matching based on a generic feature embedding ignoring the spatio-temporal consistency of the feature embedding in the training phase. In particular, our local matching strategy needs to be equipped with spatio-temporal constraints to better shape the structure of feature embedding within and across frames. Therefore, a supervised contrastive learning strategy is introduced to directly improve the consistency of query and key features for the feature-level matching and aggregation, rather than implicitly via the gradient of the decoder.

As normal, the query encoder is first adopted to map the current frame $\bm{I}_{t}$ to a 3D feature tensor, \emph{i.e.}, the query $\bm{Q}$. We then computed the object-aware contrastive loss over a non-linear projection using a light weight CNN $f_{proj}: \mathbb{R}^{HW \times D}\mapsto\mathbb{R}^{HW\times C}$ such that $\bm{Z}_{q}=f_{proj}(\bm{Q})$, as is common in contrastive learning over convolutional feature maps \cite{ChenK0H20}. The projected feature $\bm{Z}_{q}\in\mathbb{R}^{HW\times C}$ is usually called the anchor. Likewise, the key $\bm{K}$ is also projected across the index of keyframes to obtain the projected feature $\bm{Z}_{k}\in\mathbb{R}^{KHW\times C}$ through the non-linear projection head $f_{proj}$. In this work, to identify objects in each frame, we use labels downsampled to the spatial dimensions of the feature space, denoted by $\tilde{\bm{M}}_{t'}^{k'}$ ($t'\in\{1,\dots,K,t\}$ and $k'\in\{1,\dots,K'\}$), where $t'$ indexes the frames (including the current frame $t$ and the keyframes $1,\dots,K$) and $k'$ indexes the objects, with $K'$ objects in the video $\mathcal{I}$. Features projected by $f_{proj}$, in combination with $\tilde{\bm{M}}_{t'}^{k'}$, form the object-specific sets $\Phi_{k'}$ for each object $k'$.

\subsubsection{Anchor-based feature set sampling}

We sample a fixed number of features for each object $k'$ present in the anchor $\bm{Z}_{q}$, while simultaneously collecting a positive set $\mathcal{P}_{k'}$ and a negative set $\mathcal{N}_{k'}$ for each object from its instances across $K$ keyframes, \emph{i.e.}, from the projected features $\bm{Z}_{k}$. This sampling process can be described as
\begin{equation}\label{eq:anchorset}
	\Gamma\sim\{i\in\cup_{k'=1}^{K'}\Phi_{k'}\},
\end{equation}
where a fixed number of features are selected on-the-fly rather than specified as a hyperparameter, determined by the number of feature samples from the object that contains the fewest features across the video. This is motivated by the observation that small objects or regions occupy only limited spatial positions in the feature space, yet play a significant role in feature representation learning. This heuristic ensures a balanced contribution of all objects to the loss and removes the need for hyperparameter tuning. Moreover, it reduces the computational cost of each contrastive term, enabling the cross-frame contrastive learning described next.

\subsubsection{Spatio-temporal contrastive loss}

In self-supervised learning, the InfoNCE \cite{InfoNCE} loss is computed over a set of feature vectors. Similarly, we construct positive and negative sets, denoted by $\mathcal{P}_{k'}$ and $\mathcal{N}_{k'}$ ($k'\in\{1,\dots K'\}$), from a sampled anchor-based feature set as described above. In the supervised setting we consider, these sets are determined according to the object labels $\tilde{\bm{M}}_{t'}^{k'}$. Thus, without the need for dataset-specific semantic constraints, we exploit the natural occurrences of same or different object pixels across frames in the video. The cross-frames loss for the video $\mathcal{I}$ is given by
\begin{equation}\label{eq:closs}
	\mathcal{L}_{c}=\frac{1}{\left|\Gamma\right|}\sum_{i\in\Gamma}\frac{1}{\left|\mathcal{P}_{i}\right|}\sum_{j^{+}\in\mathcal{P}_{i}}\mathcal{L}(\bm{z}_{i}, \bm{z}_{j^{+}}),
\end{equation}
where $i$ is determined by Eq. \ref{eq:anchorset}. And the contrastive distance for each object is defined as
\begin{equation}\label{eq:infonce}
	\begin{gathered}
		\mathcal{L}(\bm{z}_{i}, \bm{z}_{j^{+}})=\\-\log\frac{\exp(\bm{z}_{i}\cdot\bm{z}_{j^{+}}/\tau)}{\exp(\bm{z}_{i}\cdot\bm{z}_{j^{+}}/\tau)+\sum_{j^{-}\in\mathcal{N}_{i}}^{}\exp(\bm{z}_{i}\cdot\bm{z}_{j^{-}}/\tau)},
	\end{gathered}
\end{equation}
where $\tau$ is the temperature. The choice of the positive set $\mathcal{P}$ and negative set $\mathcal{N}$ varies according to the value of $i$. For example, for object $k'$, the features include in $\Phi_{k'}$ form its positive set, while the remaining features in $\Gamma$ form its negative set.

\section{Experiments}\label{sec:exp}

\subsection{Experimental setup}

\subsubsection{Datasets}

We collected 42 X-ray angiography video sequences: 20 from the CADICA dataset \cite{CADICA} and 22 from cardiac resynchronization therapy procedures performed at two hospitals. These videos capture the injection of the contrast agent and its flow through the coronary arteries or the coronary sinus along the surface of the heart. Based on these data, we established the MOSXAV dataset, where each video contains 33$\sim$70 frames at a resolution of 512$\times$512. Experienced radiologists annotated the vascular regions, focusing on one or two frames in which the contrast agent is most prominent for each video. The training and validation sets comprise 50 sequences (2,335 frames) with dense annotations every five frames, while the test set consists of 12 sequences (488 frames) with annotations for all frames.

In addition, we conducted comprehensive experiments on two publicly available benchmark datasets: XACV \cite{XACV} and CADICA \cite{CADICA}. The XACV dataset includes 111 complete coronary artery X-ray video records from 59 patients, covering the injection, propagation through the cardiac vessels, and dissipation of the contrast agent. Each video contains an average of 86 high-resolution (512$\times$512) frames, with an equal distribution between the left and right coronary arteries. XACV provides annotations for both blood vessels and injection catheters; however, in our method these annotations are used solely for evaluation, not for training. The CADICA dataset comprises annotated invasive coronary angiography videos from 42 patients but does not provide ground-truth segmentations. To enable quantitative evaluation, we therefore selected several videos that capture the complete dissipation of the contrast agent in the cardiac vessels and contain multiple objects, such as catheters, balloons, and surgical devices, and manually annotated them to establish segmentation ground truth for use in the subsequent experiments.

\subsubsection{Evaluation metric}

Following common practice \cite{DAVIS2016,DAVIS2017} in video object segmentation, we adopt the standard metrics for the X-ray angiography video segmentation: region similarity $\mathcal{J}$, contour accuracy $\mathcal{F}$ and their average $\mathcal{J}\&\mathcal{F}$. To better evaluate the generalization performance of our method on the \texttt{test} set of the MOSXAV, we also report the $\mathcal{J}$ and $\mathcal{F}$ scores separately for ``seen'' and ``unseen'' categories, denoted by subscripts $s$ and $u$, respectively.

\subsubsection{Implementation details}\label{sssec:imp}

For the network architecture, we adopt ResNets \cite{ResNet} as the feature extractor, removing both the classification head and the final convolutional stage, which results in features with a stride of 16. The query encoder is based on ResNet-50, while the value encoder is based on ResNet-18. During training, we first pretrain on RGB video segmentation datasets \cite{DAVIS2016,DAVIS2017,YouTubeVOS}, and subsequently perform the main training on MOSXAV. The main training process takes approximately 21 hours on two NVIDIA RTX\texttrademark\ 6000 GPUs. For the training loss, we use bootstrapped cross-entropy \cite{MiVOS} for segmentation and the proposed spatio-temporal contrastive loss (Eq.~\ref{eq:closs}) for feature representation learning, with a loss weight ratio of 1:0.01. For optimization, we use Adam with a learning rate of $1$$\times$$10^{-5}$ and a weight decay of 0.05. Pretraining is performed for 150K iterations with a batch size of 16, followed by 150 iterations with a batch size of 16 for the main training. For data augmentation, we apply PyTorch's random horizontal flip, random resized crop with a crop size of 384, a scale range of (0.36, 1.0), and an aspect ratio range of (0.75, 1.25), as well as color jittering with brightness, contrast, saturation, and hue factors of 0.1, 0.03, 0.03, and 0, respectively. Additionally, we apply random affine transformations with rotation between [-15, 15] degrees and shearing between [-10, 10] degrees. During inference, we use a local sampling window size of $k$=15 for all datasets. The entire model is implemented in PyTorch \cite{PyTorch}.

\subsection{Main Results}

\begin{table*}[!t]
	\centering
	\captionsetup{font=small}
	\caption{Quantitative results on the CADICA, XACV and the \texttt{val} and \texttt{test} sets of the MOSXAV. The best performance scores are highlighted in \textbf{bold}, while \textcolor{tablegrey}{gray} indicates models that were not trained on X-ray video datasets.}
	\label{tab:SoTA}
	\begin{threeparttable}
		\resizebox{1.0\linewidth}{!}{%
			\tablestyle{4pt}{1.02}
			\begin{tabular}{@{}r|ccc|ccc|ccc|ccccc@{}}\toprule
				\multirow{3}{*}{Methods~~~~~~~~~~~~} & \multicolumn{3}{c|}{\multirow{2}{*}{CADICA}} & \multicolumn{3}{c|}{\multirow{2}{*}{XACV}} & \multicolumn{8}{c}{MOSXAV}                        \\
				& \multicolumn{3}{c|}{}          & \multicolumn{3}{c|}{}        & \multicolumn{3}{c|}{\texttt{val}} & \multicolumn{5}{c}{\texttt{test}} \\
				&       $\mathcal{J}\&\mathcal{F}\uparrow$      &   $\mathcal{J}\uparrow$       &  $\mathcal{F}\uparrow$ &       $\mathcal{J}\&\mathcal{F}\uparrow$      &   $\mathcal{J}\uparrow$       &  $\mathcal{F}\uparrow$ &$\mathcal{J}\&\mathcal{F}\uparrow$   & $\mathcal{J}\uparrow$   & $\mathcal{F}\uparrow$   &  $\mathcal{J}\&\mathcal{F}\uparrow$    & $\mathcal{J}_{s}\uparrow$   & $\mathcal{F}_{s}\uparrow$   & $\mathcal{J}_{u}\uparrow$   & $\mathcal{F}_{u}\uparrow$   \\ \midrule
				Eiseg-EdgeFlow \textcolor{tablegrey}{\scriptsize [ICCVW'2021]} \cite{EdgeFlow}        & \textcolor{tablegrey}{75.0}   &  \textcolor{tablegrey}{65.8} &      \textcolor{tablegrey}{84.3}    & \textcolor{tablegrey}{78.5} & \textcolor{tablegrey}{67.8} & \textcolor{tablegrey}{89.2} &    \textcolor{tablegrey}{74.4}        &    \textcolor{tablegrey}{66.0} &  \textcolor{tablegrey}{82.7}   &  \textcolor{tablegrey}{57.8}   &  \textcolor{tablegrey}{51.7}   &     \textcolor{tablegrey}{72.3}    &  \textcolor{tablegrey}{44.1}   &   \textcolor{tablegrey}{63.2}       \\
				Eiseg-ChestXray \textcolor{tablegrey}{\scriptsize [arXiv'22]} \cite{EISeg}       &  \textcolor{tablegrey}{73.7}  &  \textcolor{tablegrey}{64.3} &          \textcolor{tablegrey}{83.2}  & \textcolor{tablegrey}{76.1} & \textcolor{tablegrey}{65.4} & \textcolor{tablegrey}{86.8} &   \textcolor{tablegrey}{73.6}           &   \textcolor{tablegrey}{65.1}  &  \textcolor{tablegrey}{82.2}   &  \textcolor{tablegrey}{54.1}   &  \textcolor{tablegrey}{49.6}   &      \textcolor{tablegrey}{68.5}   & \textcolor{tablegrey}{40.9}    &   \textcolor{tablegrey}{57.4}       \\
				PerSAM$^{*\dag}$ \textcolor{tablegrey}{\scriptsize [arXiv'23]} \cite{PerSAM}  & \textcolor{tablegrey}{21.1}   & \textcolor{tablegrey}{17.6}  &  \textcolor{tablegrey}{24.5}       & \textcolor{tablegrey}{25.8} & \textcolor{tablegrey}{21.3} &   \textcolor{tablegrey}{30.3}     &     \textcolor{tablegrey}{20.1}       &  \textcolor{tablegrey}{16.3}   &  \textcolor{tablegrey}{24.0}   &  \textcolor{tablegrey}{6.5}   &   \textcolor{tablegrey}{3.2}  &  \textcolor{tablegrey}{4.1}       &  \textcolor{tablegrey}{8.5}   &   \textcolor{tablegrey}{10.2}       \\
				Matcher$^{*\dag}$ \textcolor{tablegrey}{\scriptsize [ICLR'24]} \cite{Matcher}    &  \textcolor{tablegrey}{36.0}  &  \textcolor{tablegrey}{27.3} &      \textcolor{tablegrey}{44.7}  & \textcolor{tablegrey}{40.2} & \textcolor{tablegrey}{30.5} &   \textcolor{tablegrey}{49.9}   &   \textcolor{tablegrey}{32.5}          & \textcolor{tablegrey}{23.9}    &   \textcolor{tablegrey}{41.1}  &   \textcolor{tablegrey}{30.4}  &   \textcolor{tablegrey}{26.2}  &    \textcolor{tablegrey}{43.5}     &  \textcolor{tablegrey}{15.2}   &      \textcolor{tablegrey}{36.7}    \\\midrule
				OSVOS \textcolor{tablegrey}{\scriptsize [CVPR'17]} \cite{OSVOS}      & 72.3   & 59.7  &   84.9     & 74.3 & 64.6 &  84.0    &   71.1     &  60.2   &  82.1   &  47.7   &  37.0   &     62.2    &   37.2  &    54.4      \\
				STM \textcolor{tablegrey}{\scriptsize [ICCV'19]} \cite{STM}         &  79.5  &  70.4 &    88.6   & 85.3 & 79.8 & 90.7  &    79.7         &  70.7   &  87.6   &   73.1  &  60.9   &   82.1      &   64.1 &    85.0      \\
				STCN \textcolor{tablegrey}{\scriptsize [NeurIPS'21]} \cite{STCN}     &  81.4  & 72.3  &   90.4     & 86.9 & 81.2 &   92.7    &     81.1      &   72.8  &  89.4   &  73.5   &  61.2   &   82.7      &  65.0   &  85.7        \\
				XMem \textcolor{tablegrey}{\scriptsize [ECCV'22]} \cite{XMem}          &  81.9  & 72.9  &   91.0   & 88.3 & 82.8 &  93.8   &   81.5           &  73.1   &  89.8   &  74.1   &  63.0   &    83.4     &   64.2  &    84.4      \\
				PerSAM-F$^{\dag}$ \textcolor{tablegrey}{\scriptsize [arXiv'23]} \cite{PerSAM} &   45.0 & 35.2  &  54.8        &  48.2  & 37.9 &  58.5      &       39.7     &  29.4   &  50.0   &  48.1   &  25.3   &    39.3     &  57.9   &   69.9       \\
				RMem \textcolor{tablegrey}{\scriptsize [CVPR'24]} \cite{RMem} & 80.9   & 71.5  &    90.3      &  85.4  & 79.1 &   91.6     &     79.5       &  70.8   &   88.2  &  73.3   &  61.0  &   82.1     & 64.5 & 85.5  \\
				LiVOS \textcolor{tablegrey}{\scriptsize [CVPR'25]} \cite{LiVOS} &  80.7  &  71.5 &   89.8       &  86.2  & 81.4 &   90.9     &     80.4       &  72.0  &  88.7   &  74.0   &   61.0  &     82.2    &  65.9   & 86.7  \\\midrule
				Ours     &  \textbf{85.0}  & \textbf{75.3}  &     \textbf{94.7}     & \textbf{89.1} & \textbf{83.7} &   \textbf{94.4}   &      \textbf{83.5}              &      \textbf{74.6}       &  \textbf{92.3}   &   \textbf{76.8}  &  \textbf{64.8}   &    \textbf{85.5}     &  \textbf{67.7}   &    \textbf{88.4}     \\ \bottomrule
			\end{tabular}%
		}
		\begin{tablenotes}
			\scriptsize
			\item $*$ indicates the training-free method. $\dag$ indicates the method using SAM.
			\item OSVOS and PerSAM-F are evaluated on the \texttt{test} set after online-training on the \texttt{test} set.
		\end{tablenotes}
	\end{threeparttable}
\end{table*}

To comprehensively evaluate the superiority of our proposed method, we compare it with state-of-the-art few-shot video segmentation methods on the CADICA and XACV datasets, as well as on the \texttt{val} and \texttt{test} sets of MOSXAV (Table \ref{tab:SoTA}). Our method is benchmarked against eight publicly available approaches from the computer vision community. For methods originally designed for single-object segmentation, including OSVOS \cite{OSVOS} and Matcher \cite{Matcher}, we extend them to the multi-object setting by splitting multiple objects into separate videos, each containing a single object.

\subsubsection{CADICA dataset}

We compare our method with top-performing FSVOS approaches on the public CADICA dataset using our provided segmentation ground truth. The detailed results are presented in the CADICA column of Table \ref{tab:SoTA}. Our method achieves substantial performance gains over existing video segmentation methods, notably outperforming the baseline FSVOS models, \emph{i.e.}, RMem, STCN, and XMem by \textbf{+4.1\%}, \textbf{+3.6\%}, and \textbf{+3.1\%} in terms of $\mathcal{J}\&\mathcal{F}$, respectively. SAM-based methods such as PerSAM and Matcher perform significantly worse, primarily due to their lack of adaptation to the X-ray image domain. By contrast, the Eiseg-series methods, \emph{i.e.}, Eiseg-EdgeFlow and Eiseg-ChestXray, exhibit better generalization to the X-ray image domain, since these matching-based approaches establish pixel correspondences based on pixel-level similarity, in a manner similar to STCN.

\begin{figure}[!h]
	\begin{center}
		\includegraphics[width=1.0\columnwidth]{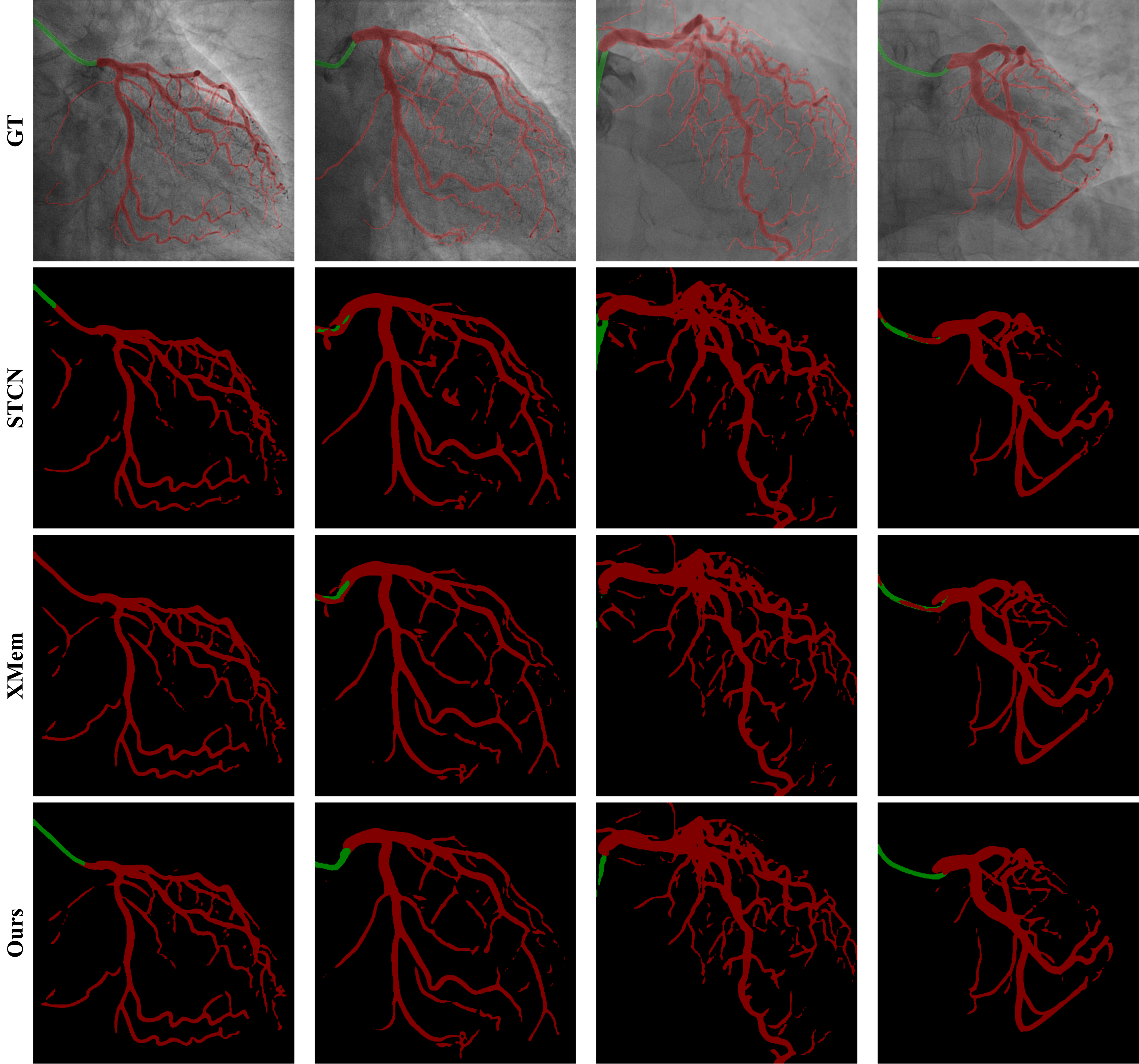}
	\end{center}
	\captionsetup{font=small}
	\caption{Qualitative results on four challenging sequences from the XACV dataset. STCN and XMem, which rely on global matching strategies, struggle with objects represented by only a small number of pixels due to the imbalanced pixel distribution problem, whereas our method performs well on these sequences. The first row presents the ground-truth masks for different objects.}
	\label{fig:qual_res_xacv}
\end{figure}

\subsubsection{XACV dataset}

We also report performance on the high-resolution XACV dataset in Table \ref{tab:SoTA}. Unlike the CADICA and MOSXAV datasets, segmentation results on XACV are evaluated without any additional training. Nevertheless, our method outperforms other FSVOS approaches, \emph{i.e.}, LiVOS, STCN, and XMem, by \textbf{2.9\%}, \textbf{2.2\%}, and \textbf{0.8\%}, respectively, in terms of $\mathcal{J}\&\mathcal{F}$. These results highlight two key observations: (1) the XACV dataset provides annotations for two frames in which the contrast agent is most prominent, allowing the support frame to capture the complete coronary artery structure and thereby facilitating accurate segmentation of subsequent frames; and (2) our MOSXAV dataset offers high resolution, fine-grained annotations and sufficient content diversity, covering multiple angiographic instances, which makes it a reliable resource for training X-ray image segmentation models.

\subsubsection{MOSXAV \texttt{val} and \texttt{test} sets}

In the MOSXAV dataset, all methods are evaluated under the same conditions, where the support frame is selected from the initial injection phase of the contrast agent and does not necessarily contain the complete coronary artery structure. This scenario is more challenging and more closely reflects real-world clinical practice. Importantly, the \texttt{test} set of MOSXAV contains unseen object classes in the \texttt{train} set, allowing for evaluation of the model's generalizability.

As shown in Table~\ref{tab:SoTA} (MOSXAV column), our method achieves the best performance among all video segmentation approaches on this challenging multi-object dataset. On the \texttt{val} set, our method attains an $\mathcal{J}\&\mathcal{F}$ score of \textbf{83.5\%}, outperforming the second-best method (XMem) and the third-best method (STCN) by \textbf{2.0\%} and \textbf{2.4\%}, respectively. On the \texttt{test} set, particularly for unseen object classes, our approach demonstrates consistent performance gains over XMem and LiVOS, improving the $\mathcal{J}\&\mathcal{F}$ score from 74.1\% to 76.8\% and from 74.0\% to 76.8\%, respectively.

\subsubsection{Qualitative results}

\begin{figure*}[!tb]
	\begin{center}
		\includegraphics[width=1.0\linewidth]{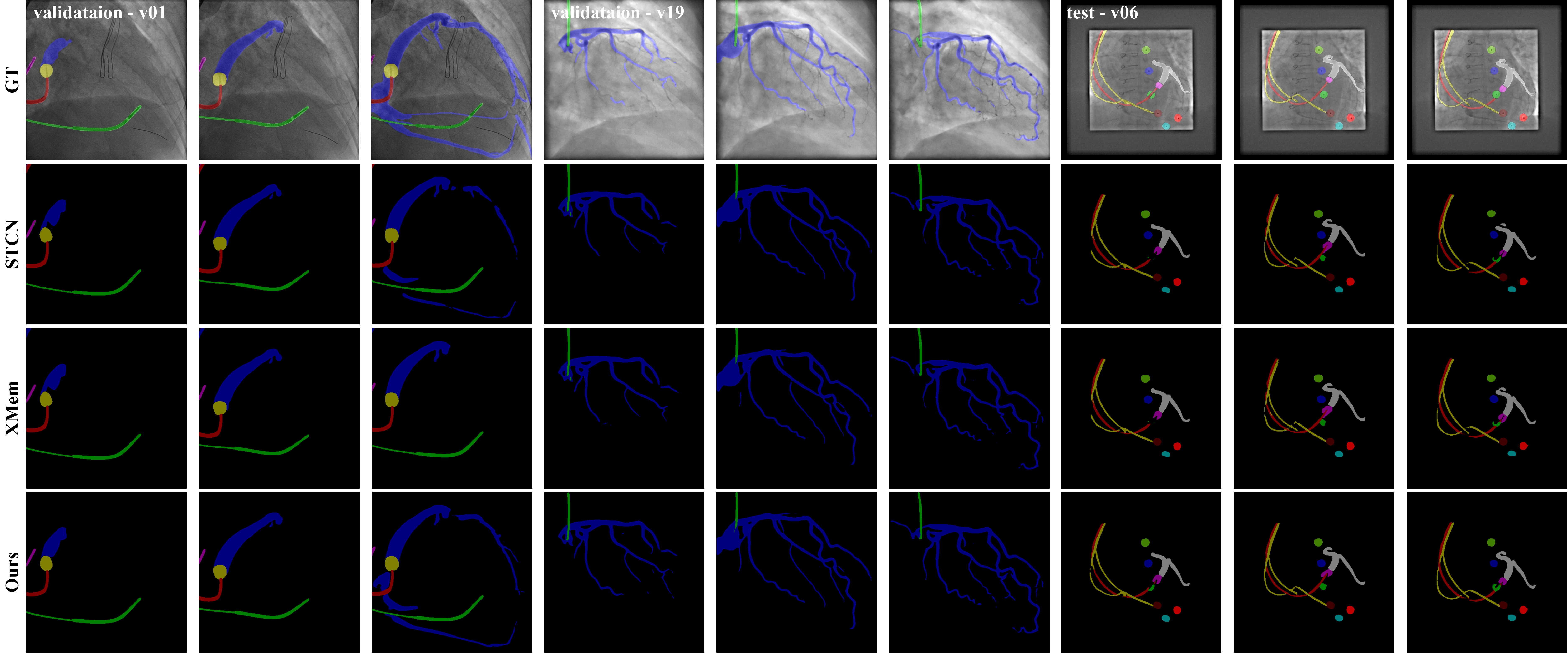}
	\end{center}
	\captionsetup{font=small}
	\caption{Qualitative results on three challenging sequences from the \texttt{val} and \texttt{test} sets of MOSXAV. The two validation sequences (\emph{i.e.}, \textit{v01} and \textit{v19}) present complex scenarios with overlapping vessels and severe cardiac motion. The \texttt{test} sequence (\emph{i.e.}, \textit{v06}) contains unseen object classes not present in the training set, specifically six spherical objects, and was captured using low-dose fluoroscopy. The first row shows the ground truth masks for the different objects.}
	\label{fig:qual_res_mosxav}
\end{figure*}

Figure \ref{fig:qual_res_xacv} provides a qualitative comparison of our method against STCN and XMem on representative examples from the XACV dataset. We observe that our approach effectively handles diverse and challenging scenarios, producing more accurate results. Figure \ref{fig:qual_res_mosxav} presents a qualitative comparison on the \texttt{val} and \texttt{test} sets of MOSXAV. Compared with state-of-the-art FSVOS methods, our approach achieves more stable and accurate mask-tracking results, even in challenging scenarios involving overlapping vessels (\textit{v01}), severe cardiac motion (\textit{v19}) and unseen object classes under low-dose fluoroscopy (\textit{v06}).

\subsection{Ablation Study}

To demonstrate the effectiveness of the local sampling module in our method, we perform an ablation study on the \texttt{val} set of MOSXAV. The evaluation criterion is the mean region similarity ($\mathcal{J}$) and frames par second (FPS).

\subsubsection{Training objective} 

We investigate our overall training objective, as described in Section \ref{sssec:imp}, which consists of the cross-entropy loss $\mathcal{L}_{ce}$ and the spatio-temporal contrastive loss $\mathcal{L}_{c}$. As shown in Table \ref{tab:abla:loss}, the model trained with $\mathcal{L}_{ce}$ alone achieves a mean $\mathcal{J}$ score of 73.8\%. Incorporating $\mathcal{L}_{c}$ yields an additional improvement of 0.8\%, highlighting the benefit of explicitly shaping the feature representation.

\begin{table}[]
	\centering
	\tiny
	\captionsetup{font=small}
	\caption{Ablation study of training objective on the MOSXAV dataset, measured by mean $\mathcal{J}$.}
	\label{tab:abla:loss}
	\tablestyle{15pt}{1.05}
	\resizebox{0.28\columnwidth}{!}{
		\begin{tabular}{@{\hskip 4pt}cc|c@{\hskip 4pt}}
			\toprule
			$\mathcal{L}_{ce}$& $\mathcal{L}_{c}$  & Mean $\mathcal{J}\uparrow$ \\ \midrule
			\cmark &  & 73.8 \\
			\cmark & \cmark & 74.6 \textcolor{tablegreen}{\scriptsize ($\uparrow$0.8)} \\ \bottomrule
		\end{tabular}
	}
\end{table}

\begin{figure*}[!tb]
	\begin{center}
		\includegraphics[width=1.0\linewidth]{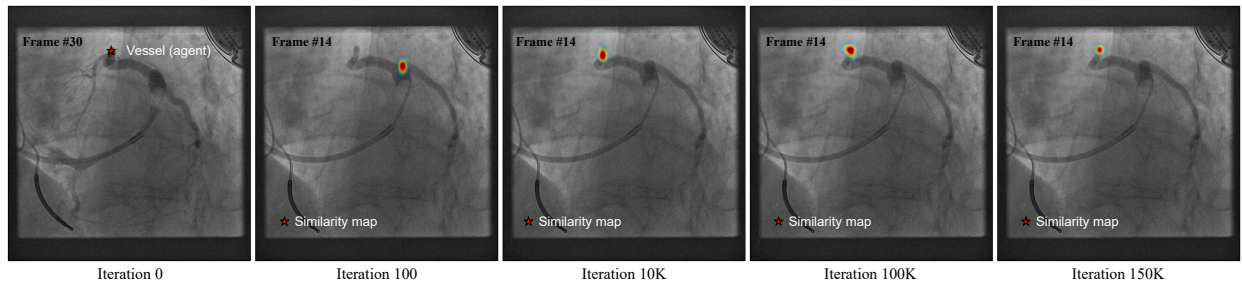}
	\end{center}
	\captionsetup{font=small}
	\caption{Visualization of the pixels in the keyframe with the highest similarity scores relative to the selected coronary vessel pixel (red star - \tikzsymbol[star]{fill=red,star point ratio=2.618}{1pt}) in the query frame during training. Based on the similarity scores between the selected pixel in the query frame and the most relevant pixels in keyframe, the model gradually learns to focus on the vessel structure while suppressing responses to irrelevant distractors.}
	\label{fig:training_process}
\end{figure*}

\begin{table*}[h]
	\centering
	\captionsetup{font=small}
	\caption{Ablation studies on the MOSXAV, measured by mean $\mathcal{J}$ and FPS.}
	\label{tab:abla}
	\begin{threeparttable}
	\resizebox{1.0\textwidth}{!}{
		\subfloat[{Sampling operation function} \label{table:abla:sampling_op}]{
			\tablestyle{10pt}{1.05}
			\begin{tabular}{@{\hskip 4pt}c|cccc@{\hskip 4pt}}
				\toprule
				\multirow{2}{*}{Sampling Strategies} & \multirow{2}{*}{Mean $\mathcal{J}\uparrow$} & \multicolumn{3}{c}{FPS $\uparrow$} \\ \cmidrule{3-5}
				&                     & RTX\texttrademark\ 6000 GPU & i9-13900H CPU & R9 7940HS CPU \\\midrule
				baseline &    72.8  & 20 & 0.289$\pm$0.006 & 0.283$\pm$0.004 \\
				feature shift&    72.4 \textcolor{tablered}{\scriptsize ($\downarrow$0.4)}  & 10 \textcolor{tablered}{\scriptsize ($\downarrow$10)}& 0.268$\pm$0.003 \textcolor{tablered}{\scriptsize ($\downarrow$0.021)} & 0.257$\pm$0.002 \textcolor{tablered}{\scriptsize ($\downarrow$0.026)}\\
				depth-wise conv&    73.3 \textcolor{tablegreen}{\scriptsize ($\uparrow$0.5)}   & 19 \textcolor{tablered}{\scriptsize ($\downarrow$1)}& 0.282$\pm$0.001 \textcolor{tablered}{\scriptsize ($\downarrow$0.007)}& 0.275$\pm$0.005 \textcolor{tablered}{\scriptsize ($\downarrow$0.008)}\\
				deformable conv&    73.8 \textcolor{tablegreen}{\scriptsize ($\uparrow$1.0)}  & 16 \textcolor{tablered}{\scriptsize ($\downarrow$4)}& - & - \\
				2D neighborhood attention&    74.5 \textcolor{tablegreen}{\scriptsize ($\uparrow$1.7)}  & 22 \textcolor{tablegreen}{\scriptsize ($\uparrow$2)}& - & - \\
				unfold (\textbf{ours}) &    \textbf{74.6} \textcolor{tablegreen}{\scriptsize ($\uparrow$1.8)}  & \textbf{25} \textcolor{tablegreen}{\scriptsize ($\uparrow$5)}& 0.301$\pm$0.004 \textcolor{tablegreen}{\scriptsize ($\uparrow$0.012)} & 0.293$\pm$0.001 \textcolor{tablegreen}{\scriptsize ($\uparrow$0.010)}\\\bottomrule
			\end{tabular}%
		}\hfill\hspace{5pt}
		\subfloat[{Neighborhood region}\label{table:abla:size}]{%
			\tablestyle{10pt}{1.05}
			\begin{tabular}{@{\hskip 4pt}c|cc@{\hskip 4pt}}
				\toprule
				Sampling Size $k$ & Mean $\mathcal{J}\uparrow$ & FPS $\uparrow$ \\ \cmidrule[0pt]{1-1} \midrule
				$k$=5&     73.1 & \textbf{43}\\
				$k$=7&     73.3  & 37\\
				$k$=9&     74.5  & 32\\
				$k$=13&     74.5 &  27\\
				$k$=15&     \textbf{74.6} & 25\\
				$k$=17&     74.5 & 22\\
				$k$=19&     74.3 & 19\\\bottomrule
			\end{tabular}%
		}\hfill\hspace{5pt}
		\subfloat[{Keyframe sampling interval}\label{table:abla:mem}]{%
			\tablestyle{10pt}{1.05}
			\begin{tabular}{@{\hskip 4pt}c|cc@{\hskip 4pt}}
				\toprule
				$r$-frames & Mean $\mathcal{J}\uparrow$ & FPS $\uparrow$ \\ \cmidrule[0pt]{1-1} \midrule 
				$r$=1&     74.2  & 10 \\
				$r$=2&     74.7  & 16 \\
				$r$=3&     74.2 &  20 \\
				$r$=5&     73.7 & 23 \\
				$r$=6&     \textbf{74.6} & 25 \\
				$r$=8&     74.5 & 27 \\
				$r$=10&     73.6 & \textbf{28} \\\bottomrule
			\end{tabular}%
		}\hfill
	}
	\begin{tablenotes}
			\scriptsize
			\item \parbox{\textwidth}{- denotes that the operation is exclusively optimized for GPU via specialized CUDA kernels and lacks a corresponding CPU implementation.}
		\end{tablenotes}
	\end{threeparttable}
\end{table*}

\subsubsection{Sampling strategies}

To evaluate the efficacy of the local sampling module $\phi(\cdot;\cdot)$ in Eq. \ref{eq:unfold}, we systematically replace it with alternative operations, specifically the feature shifting mechanism (\emph{i.e.}, spatial feature slicing) utilized in Window Attention \cite{Swin} and the depth-wise convolution adopted in Slide Attention \cite{SlideTransformer}. For a fair comparison, these alternatives are implemented using the official PyTorch implementations from their respective open-source repositories. We adopt the global feature matching strategy of STCN \cite{STCN} as our baseline. Furthermore, we implement the local sampling process using specialized CUDA kernels (\emph{e.g.}, deformable convolution \cite{DCN} and neighborhood attention \cite{NAT}) and benchmark their performance on an RTX\texttrademark\ 6000 GPU. Additionally, we assess the CPU inference efficiency of the local sampling module using Intel\textsuperscript{\textregistered}\ Core\texttrademark\ i9-13900H and AMD Ryzen\texttrademark\ 9 7940HS processors. Results are summarized in Table \ref{table:abla:sampling_op}. For CPU inference speed (FPS), each implementation is executed ten times under identical hyperparameters settings, awith the mean and standard deviation reported.

Compared with the baseline, our local sampling module implemented via the unfold sampling function (Eq. \ref{eq:unfold}) improves segmentation performance by \textbf{1.8\%} in terms of mean $\mathcal{J}$. More importantly, it increases inference speed to \textbf{25} FPS on the GPU (\emph{vs.} 20 FPS for the baseline) and to \textbf{0.301} FPS and \textbf{0.293} FPS on the Intel\textsuperscript{\textregistered}\ Core\texttrademark\ i9-13900H CPU and the AMD Ryzen\texttrademark\ 9 7940HS CPU, respectively (\emph{vs.} 0.289 and 0.283 for the baseline). To improve efficiency, we first partition the image into patches, similar to the patch embedding module in ViT \cite{ViT}, and then compute similarity within each local window. The main motivation for this design is that applying im2col-like local sampling individually to every pixel is prohibitively expensive. Regarding clinical utility, our current PyTorch implementation requires approximately 3.32 and 3.41 seconds per image on the aforementioned Intel and AMD CPUs. While this execution speed is not yet optimized for real-time intraoperative use, it is well-suited for offline analysis and automated batch processing within the research phase. Our method achieves a performance gain over the global matching baseline while maintaining competitive accuracy, suggesting it can be effectively integrated into clinical workflows that prioritize reliability over latency, such as retrospective cohort studies or pre-procedural planning. Consistent with this observation, we found that the feature-shift operation based on im2col-like functions yield both lower segmentation accuracy and higher computational costs.  For fixed and deformable convolution-based sampling, the learnable bias terms fail to capture effective inductive biases due to the depth-wise nature of these operations. In our implementation, we fix the kernel weights to 1 and optimize only the bias parameters. Compared with the baseline, these two sampling operations improve accuracy by 0.5\% and 1.0\%, respectively, both resulted in reduced inference speeds across GPU and CPU architectures. In contrast, while 2D neighborhood attention achieves strong performance and efficiency, its deployment, much like deformable convolution, remains constrained by a dependency on device-specific CUDA implementations.

Figure \ref{fig:training_process} visualizes the evolution of the learned correspondences process, which is crucial for understanding how feature representations develop during training. In the early iterations, the model fails to capture the relevant regions due to the presence of visually similar distractors and textured backgrounds (\emph{e.g.}, the similarity map at 100 iterations). As training progresses, the most relevant pixels gradually concentrate on the vessel structures, while responses to irrelevant distractors are increasingly suppressed. These qualitative results demonstrate that our local sampling strategy effectively captures meaningful contextual information. Notably, even at 10K iterations, the model is able to identify most relevant of the vessel structure, indicating that the proposed local sampling mechanism enables efficient learning of discriminative features.

\subsubsection{Sampling size}

Table \ref{table:abla:size} reports the performance of our approach with respect to the sampling region size $k$. As $k$ increases, the mean $\mathcal{J}$ first improves and then declines. Notably, using a larger region ($k$=9) yields a clear performance gain (73.1\%$\rightarrow$74.5\%). The score further improves at $k=15$; however, increasing $k$ beyond 15 results in a slight drop in performance. Based on this observation, we empirically set $k=15$, which provides the best trade-off between accuracy and computational cost. Smaller sampling sizes achieve higher FPS, as the matrix multiplication on $\mathcal{M}=\{(i, j),c_{ij}\}$ in Section \ref{sssec:matching} involves fewer operations.

\subsubsection{Keyframes sampling}

We compare different keyframe sampling intervals $r$ in Table \ref{table:abla:mem}. For $r=1$, every frame in the video is selected as a support frame and stored in the external memory to build the feature memory, yielding a baseline score of 74.2\%. As $r$ increases, the FPS improves due to the reduced size of the correspondence matrix $\bm{C}$ in Section \ref{sssec:matching}, while the mean $\mathcal{J}$ initially rises and then drops. Based on this observation, we empirically set $r=6$ to achieve a better trade-off between accuracy and computational cost.

\section{Conclusion}\label{sec:con}

In this paper, we propose an FSVOS method for angiography video segmentation, aiming to minimize the prohibitive costs of expert annotation. By focusing on localized neighborhood regions, our method achieves superior efficiency and generalizability in feature matching. To ensure broad applicability across diverse hospital computing infrastructures, we replaced inefficient standard im2col-like implementations and hardware-dependent kernels with a non-parametric local sampling operation. Extensive evaluations on the CADICA, XACV, and MOSXAV datasets demonstrate that our approach outperforms state-of-the-art video segmentation methods, providing the precise vessel delineation required for clinical decision-making.

Despite these methodological advances, translating a retrospective research prototype into a real-world clinical application presents significant challenges, particularly regarding data privacy, system interoperability, regulatory compliance, and real-time deployment. To bridge this gap, our future work will focus on three strategic fronts: (1) Clinical prototyping: We aim to develop a DICOM-compatible plugin via open-source platforms such as 3D Slicer, facilitating intuitive clinician interaction and seamless data integration. (2) Regulatory and security standards: We will further optimize inference latency and strengthen data security protocols to align with stringent medical device regulatory standards. (3) High-performance deployment: To achieve real-time capability, we will transition from our current Python-based implementation to high-efficiency C++ frameworks, specifically leveraging LibTorch and ONNX Runtime to enable advanced parallel processing and hardware acceleration. We believe this work establishes a robust foundation for ``human-in-the-loop'' AI systems, ultimately enhancing standard diagnostic workflows with reliable, high-performance assistance.

\section*{Acknowledgements}

This work was supported by EPSRC UK (EP/X023826/1).

\bibliographystyle{unsrt}  
\bibliography{references}

\begin{thebibliography}{10}

\bibitem{OSVOS}
S.~Caelles, K.-K. Maninis, J.~Pont-Tuset, L.~Leal-Taix{\'e}, D.~Cremers, and
  L.~Van~Gool.
\newblock One-shot video object segmentation.
\newblock In {\em Proceedings of the IEEE Conference on Computer Vision and
  Pattern Recognition (CVPR)}, pages 5320--5329, 2017.

\bibitem{MaskTrack}
Federico Perazzi, Anna Khoreva, Rodrigo Benenson, Bernt Schiele, and Alexander
  Sorkine-Hornung.
\newblock Learning video object segmentation from static images.
\newblock In {\em Proceedings of the IEEE Conference on Computer Vision and
  Pattern Recognition (CVPR)}, pages 3491--3500, 2017.

\bibitem{STM}
Seoung~Wug Oh, Joon-Young Lee, Ning Xu, and Seon~Joo Kim.
\newblock Video object segmentation using space-time memory networks.
\newblock In {\em Proceedings of the IEEE International Conference on Computer
  Vision (ICCV)}, pages 9225--9234, 2019.

\bibitem{STCN}
Ho~Kei Cheng, Yu-Wing Tai, and Chi-Keung Tang.
\newblock Rethinking space-time networks with improved memory coverage for
  efficient video object segmentation.
\newblock In {\em Advances in Neural Information Processing Systems (NeurIPS)},
  pages 11781--11794, 2021.

\bibitem{XMem}
Ho~Kei Cheng and Alexander~G Schwing.
\newblock {XMem}: Long-term video object segmentation with an atkinson-shiffrin
  memory model.
\newblock In {\em Proceedings of the European Conference on Computer Vision
  (ECCV)}, volume 13688, pages 640--658, 2022.

\bibitem{RAB}
Zixuan Zheng, Yilei Shi, Chunlei Li, Jingliang Hu, Xiao~Xiang Zhu, and Lichao
  Mou.
\newblock Reducing annotation burden: Exploiting image knowledge for few-shot
  medical video object segmentation via spatiotemporal consistency relearning.
\newblock In {\em Proceedings of Medical Image Computing and Computer-Assisted
  Intervention (MICCAI)}, pages 272 -- 282, 2024.

\bibitem{CENet}
Zaiwang Gu, Jun Cheng, Huazhu Fu, Kang Zhou, Huaying Hao, Yitian Zhao, Tianyang
  Zhang, Shenghua Gao, and Jiang Liu.
\newblock Ce-net: Context encoder network for 2d medical image segmentation.
\newblock {\em IEEE Transactions on Medical Imaging}, 38(10):2281--2292, 2019.

\bibitem{AOT}
Zongxin Yang, Yunchao Wei, and Yi~Yang.
\newblock Associating objects with transformers for video object segmentation.
\newblock In {\em Advances in Neural Information Processing Systems (NeurIPS)},
  pages 2491--2502, 2021.

\bibitem{SASA}
Prajit Ramachandran, Niki Parmar, Ashish Vaswani, Irwan Bello, Anselm Levskaya,
  and Jon Shlens.
\newblock Stand-alone self-attention in vision models.
\newblock In {\em Advances in Neural Information Processing Systems (NeurIPS)},
  volume~32, 2019.

\bibitem{Swin}
Ze~Liu, Yutong Lin, Yue Cao, Han Hu, Yixuan Wei, Zheng Zhang, Stephen Lin, and
  Baining Guo.
\newblock Swin transformer: {Hierarchical} vision transformer using shifted
  windows.
\newblock In {\em Proceedings of the IEEE International Conference on Computer
  Vision (ICCV)}, pages 9992--10002, 2021.

\bibitem{Focal}
Jianwei Yang, Chunyuan Li, Pengchuan Zhang, Xiyang Dai, Bin Xiao, Lu~Yuan, and
  Jianfeng Gao.
\newblock Focal attention for long-range interactions in vision transformers.
\newblock In {\em Advances in Neural Information Processing Systems (NeurIPS)},
  volume~34, pages 30008--30022, 2021.

\bibitem{NAT}
Ali Hassani, Steven Walton, Jiachen Li, Shen Li, and Humphrey Shi.
\newblock Neighborhood attention transformer.
\newblock In {\em Proceedings of the IEEE/CVF Conference on Computer Vision and
  Pattern Recognition (CVPR)}, pages 6185--6194, June 2023.

\bibitem{SlideTransformer}
Xuran Pan, Tianzhu Ye, Zhuofan Xia, Shiji Song, and Gao Huang.
\newblock Slide-transformer: Hierarchical vision transformer with local
  self-attention.
\newblock In {\em Proceedings of the IEEE/CVF Conference on Computer Vision and
  Pattern Recognition (CVPR)}, pages 2082--2091, 2023.

\bibitem{TII2025Zhang}
Pengfei Zhang, Hong Sun, Zhikun Zhang, Xiang Cheng, Youwen Zhu, and Ji~Zhang.
\newblock Privacy-preserving recommendations with mixture model-based matrix
  factorization under local differential privacy.
\newblock {\em IEEE Transactions on Industrial Informatics}, 21(7):5451--5459,
  2025.

\bibitem{LiVOS}
Qin Liu, Jianfeng Wang, Zhengyuan Yang, Linjie Li, Kevin Lin, Marc Niethammer,
  and Lijuan Wang.
\newblock Livos: Light video object segmentation with gated linear matching.
\newblock In {\em Proceedings of the IEEE Conference on Computer Vision and
  Pattern Recognition (CVPR)}, pages 8668--8678, 2025.

\bibitem{CADICA}
Ariadna Jim{\'e}nez-Partinen, Miguel~A Molina-Cabello, Karl Thurnhofer-Hemsi,
  Esteban~J Palomo, Jorge Rodr{\'\i}guez-Capit{\'a}n, Ana~I Molina-Ramos, and
  Manuel Jim{\'e}nez-Navarro.
\newblock Cadica: A new dataset for coronary artery disease detection by using
  invasive coronary angiography.
\newblock {\em Expert Systems}, 41(12):e13708, 2024.

\bibitem{XACV}
Chun-Hung Wu, Shih-Hong Chen, Chih-Yao Hu, Hsin-Yu Wu, Kai-Hsin Chen, Yu-You
  Chen, Chih-Hai Su, Chih-Kuo Lee, and Yu-Lun Liu.
\newblock Denver: Deformable neural vessel representations for unsupervised
  video vessel segmentation.
\newblock In {\em Proceedings of the IEEE/CVF Conference on Computer Vision and
  Pattern Recognition (CVPR)}, pages 15682--15692, 2025.

\bibitem{UNet}
Olaf Ronneberger et~al.
\newblock U-net: Convolutional networks for biomedical image segmentation.
\newblock In {\em Medical Image Computing and Computer-Assisted Intervention
  (MICCAI)}, pages 234--241, 2015.

\bibitem{T-Net}
Tae~Joon Jun, Jihoon Kweon, Young-Hak Kim, and Daeyoung Kim.
\newblock T-net: Nested encoder-decoder architecture for the main vessel
  segmentation in coronary angiography.
\newblock {\em Neural Networks}, 128:216--233, 2020.

\bibitem{SVSNet}
Dongdong Hao, Song Ding, Linwei Qiu, Yisong Lv, Baowei Fei, Yueqi Zhu, and
  Binjie Qin.
\newblock Sequential vessel segmentation via deep channel attention network.
\newblock {\em Neural Networks}, 128:172--187, 2020.

\bibitem{KHAN}
Tariq~M. Khan, Syed~S. Naqvi, Antonio Robles-Kelly, and Imran Razzak.
\newblock Retinal vessel segmentation via a multi-resolution contextual network
  and adversarial learning.
\newblock {\em Neural Networks}, 165:310--320, 2023.

\bibitem{ViT}
Alexey Dosovitskiy, Lucas Beyer, Alexander Kolesnikov, Dirk Weissenborn,
  Xiaohua Zhai, Thomas Unterthiner, Mostafa Dehghani, Matthias Minderer, Georg
  Heigold, Sylvain Gelly, et~al.
\newblock An image is worth 16x16 words: Transformers for image recognition at
  scale.
\newblock In {\em International Conference on Learning Representations (ICLR)},
  2021.

\bibitem{G2ViT}
Hao Xu and Yun Wu.
\newblock G2vit: Graph neural network-guided vision transformer enhanced
  network for retinal vessel and coronary angiograph segmentation.
\newblock {\em Neural Networks}, 176:106356, 2024.

\bibitem{CBAMViT}
Yuqi Ma, Huamin Wang, Hangchi Shen, Shukai Duan, and Shiping Wen.
\newblock Analog spiking u-net integrating cbam\&vit for medical image
  segmentation.
\newblock {\em Neural Networks}, 181:106765, 2025.

\bibitem{Baur2016MIAR}
Christoph Baur et~al.
\newblock Cathnets: Detection and single-view depth prediction of catheter
  electrodes.
\newblock In {\em Medical Imaging and Augmented Reality}, pages 38--49, 2016.

\bibitem{Ma2015TMI}
Xianliang Wu et~al.
\newblock Fast catheter segmentation from echocardiographic sequences based on
  segmentation from corresponding x-ray fluoroscopy for cardiac catheterization
  interventions.
\newblock {\em IEEE Transactions on Medical Imaging}, 34(4):861--876, 2015.

\bibitem{Ambrosini2017MICCAI}
Pierre Ambrosini et~al.
\newblock Fully automatic and real-time catheter segmentation in x-ray
  fluoroscopy.
\newblock In {\em Medical Image Computing and Computer-Assisted Intervention
  (MICCAI)}, pages 577--585, 2017.

\bibitem{CathMTL}
Lin Xi, Yingliang Ma, Ethan Koland, Sandra Howell, Aldo Rinaldi, and Kawal~S
  Rhode.
\newblock Catheter detection and segmentation in x-ray images via multi-task
  learning.
\newblock {\em International Journal of Computer Assisted Radiology and
  Surgery}, pages 1--11, 2025.

\bibitem{KIM2022CMIG}
Taeouk Kim et~al.
\newblock A learning-based, region of interest-tracking algorithm for catheter
  detection in echocardiography.
\newblock {\em Computerized Medical Imaging and Graphics}, 100:102106, 2022.

\bibitem{yang2019catheter}
Hongxu Yang et~al.
\newblock Catheter localization in 3d ultrasound using voxel-of-interest-based
  convnets for cardiac intervention.
\newblock {\em International journal of computer assisted radiology and
  surgery}, 14:1069--1077, 2019.

\bibitem{Nguyen2020ICRA}
Anh Nguyen et~al.
\newblock End-to-end real-time catheter segmentation with optical flow-guided
  warping during endovascular intervention.
\newblock In {\em IEEE International Conference on Robotics and Automation
  (ICRA)}, pages 9967--9973, 2020.

\bibitem{Ranne2024ICRA}
Alex Ranne et~al.
\newblock Aiareseg: Catheter detection and segmentation in interventional
  ultrasound using transformers.
\newblock In {\em IEEE International Conference on Robotics and Automation
  (ICRA)}, pages 8187--8194, 2024.

\bibitem{COSNet}
Xiankai Lu, Wenguan Wang, Chao Ma, Jianbing Shen, Ling Shao, and Fatih Porikli.
\newblock See more, know more: {Unsupervised} video object segmentation with
  co-attention siamese networks.
\newblock In {\em Proceedings of the IEEE/CVF Conference on Computer Vision and
  Pattern Recognition (CVPR)}, pages 3618--3627, 2019.

\bibitem{IMCNet}
Lin Xi, Weihai Chen, Xingming Wu, Zhong Liu, and Zhengguo Li.
\newblock Implicit motion-compensated network for unsupervised video object
  segmentation.
\newblock {\em IEEE Transactions on Circuits and Systems for Video Technology},
  32(9):6279--6292, 2022.

\bibitem{ClusterNet}
Lin Xi, Weihai Chen, Xingming Wu, Zhong Liu, and Zhengguo Li.
\newblock Online unsupervised video object segmentation via contrastive motion
  clustering.
\newblock {\em IEEE Transactions on Circuits and Systems for Video Technology},
  34(2):995--1006, 2024.

\bibitem{OSVOSS}
K.-K. Maninis, S.~Caelles, Y.~Chen, J.~Pont-Tuset, L.~Leal-Taixé, D.~Cremers,
  and L.~Van~Gool.
\newblock Video object segmentation without temporal information.
\newblock {\em IEEE Transactions on Pattern Analysis and Machine Intelligence},
  41(6):1515--1530, 2019.

\bibitem{CFBI}
Zongxin Yang, Yunchao Wei, and Yi~Yang.
\newblock Collaborative video object segmentation by foreground-background
  integration.
\newblock In {\em Proceedings of the European Conference on Computer Vision
  (ECCV)}, volume 12350, pages 332--348, 2020.

\bibitem{FEELVOS}
Paul Voigtlaender, Yuning Chai, Florian Schroff, Hartwig Adam, Bastian Leibe,
  and Liang-Chieh Chen.
\newblock Feelvos: Fast end-to-end embedding learning for video object
  segmentation.
\newblock In {\em Proceedings of the IEEE Conference on Computer Vision and
  Pattern Recognition (CVPR)}, pages 9473--9482, 2019.

\bibitem{RVOS}
Carles Ventura, Miriam Bellver, Andreu Girbau, Amaia Salvador, Ferran Marques,
  and Xavier Giro-i Nieto.
\newblock Rvos: End-to-end recurrent network for video object segmentation.
\newblock In {\em Proceedings of the IEEE Conference on Computer Vision and
  Pattern Recognition (CVPR)}, pages 5272--5281, 2019.

\bibitem{SiamMask}
Qiang Wang, Li~Zhang, Luca Bertinetto, Weiming Hu, and Philip~H.S. Torr.
\newblock Fast online object tracking and segmentation: A unifying approach.
\newblock In {\em Proceedings of the IEEE Conference on Computer Vision and
  Pattern Recognition (CVPR)}, pages 1328--1338, 2019.

\bibitem{MASKRNN}
Yuan-Ting Hu, Jia-Bin Huang, and Alexander Schwing.
\newblock Maskrnn: Instance level video object segmentation.
\newblock {\em Advances in Neural Information Processing Systems (NeurIPS)},
  30, 2017.

\bibitem{FRTM}
Andreas Robinson, Felix~Jaremo Lawin, Martin Danelljan, Fahad~Shahbaz Khan, and
  Michael Felsberg.
\newblock Learning fast and robust target models for video object segmentation.
\newblock In {\em Proceedings of the IEEE Conference on Computer Vision and
  Pattern Recognition (CVPR)}, pages 7404--7413, 2020.

\bibitem{MiVOS}
Ho~Kei Cheng, Yu-Wing Tai, and Chi-Keung Tang.
\newblock Modular interactive video object segmentation: Interaction-to-mask,
  propagation and difference-aware fusion.
\newblock In {\em Proceedings of the IEEE Conference on Computer Vision and
  Pattern Recognition (CVPR)}, pages 5555--5564, 2021.

\bibitem{Transformer}
Ashish Vaswani, Noam Shazeer, Niki Parmar, Jakob Uszkoreit, Llion Jones,
  Aidan~N. Gomez, Lukasz Kaiser, and Illia Polosukhin.
\newblock Attention is all you need.
\newblock In {\em Advances in Neural Information Processing Systems (NeurIPS)},
  volume~30, pages 5998--6008, 2017.

\bibitem{RegionViT}
Chun{-}Fu Chen, Rameswar Panda, and Quanfu Fan.
\newblock Regionvit: Regional-to-local attention for vision transformers.
\newblock In {\em International Conference on Learning Representations (ICLR)},
  2022.

\bibitem{DCN}
Jifeng Dai, Haozhi Qi, Yuwen Xiong, Yi~Li, Guodong Zhang, Han Hu, and Yichen
  Wei.
\newblock Deformable convolutional networks.
\newblock In {\em Proceedings of the IEEE International Conference on Computer
  Vision (ICCV)}, pages 764--773, 2017.

\bibitem{DeformableDETR}
Xizhou Zhu, Weijie Su, Lewei Lu, Bin Li, Xiaogang Wang, and Jifeng Dai.
\newblock Deformable {DETR}: Deformable transformers for end-to-end object
  detection.
\newblock In {\em International Conference on Learning Representations (ICLR)},
  2021.

\bibitem{PnP-DETR}
Tao Wang, Li~Yuan, Yunpeng Chen, Jiashi Feng, and Shuicheng Yan.
\newblock Pnp-detr: Towards efficient visual analysis with transformers.
\newblock In {\em Proceedings of the IEEE International Conference on Computer
  Vision (ICCV)}, pages 4641--4650, 2021.

\bibitem{SparseDETR}
Byungseok Roh, JaeWoong Shin, Wuhyun Shin, and Saehoon Kim.
\newblock Sparse {DETR}: Efficient end-to-end object detection with learnable
  sparsity.
\newblock In {\em International Conference on Learning Representations (ICLR)},
  2022.

\bibitem{wu2018unsupervised}
Zhirong Wu, Yuanjun Xiong, Stella~X. Yu, and Dahua Lin.
\newblock Unsupervised feature learning via non-parametric instance
  discrimination.
\newblock In {\em Proceedings of the IEEE Conference on Computer Vision and
  Pattern Recognition (CVPR)}, pages 3733--3742, 2018.

\bibitem{Henaff20}
Olivier~J. H{\'{e}}naff.
\newblock Data-efficient image recognition with contrastive predictive coding.
\newblock In {\em Proceedings of the International Conference on Machine
  Learning (ICML)}, volume 119, pages 4182--4192, 2020.

\bibitem{Hjelm19}
R~Devon Hjelm, Alex Fedorov, Samuel Lavoie-Marchildon, Karan Grewal, Phil
  Bachman, Adam Trischler, and Yoshua Bengio.
\newblock Learning deep representations by mutual information estimation and
  maximization.
\newblock In {\em International Conference on Learning Representations (ICLR)},
  2019.

\bibitem{CMC}
Yonglong Tian, Dilip Krishnan, and Phillip Isola.
\newblock Contrastive multiview coding.
\newblock In {\em Proceedings of the European Conference on Computer Vision
  (ECCV)}, volume 12356, pages 776--794, 2020.

\bibitem{Sermanet18}
Pierre Sermanet, Corey Lynch, Yevgen Chebotar, Jasmine Hsu, Eric Jang, Stefan
  Schaal, Sergey Levine, and Google Brain.
\newblock Time-contrastive networks: Self-supervised learning from video.
\newblock In {\em International Conference on Robotics and Automation (ICRA)},
  pages 1134--1141, 2018.

\bibitem{ChenK0H20}
Ting Chen, Simon Kornblith, Mohammad Norouzi, and Geoffrey~E. Hinton.
\newblock A simple framework for contrastive learning of visual
  representations.
\newblock In {\em Proceedings of the International Conference on Machine
  Learning (ICML)}, volume 119, pages 1597--1607, 2020.

\bibitem{Tschannen2020On}
Michael Tschannen, Josip Djolonga, Paul~K. Rubenstein, Sylvain Gelly, and Mario
  Lucic.
\newblock On mutual information maximization for representation learning.
\newblock In {\em International Conference on Learning Representations (ICLR)},
  2020.

\bibitem{GutmannH10}
Michael Gutmann and Aapo Hyv{\"{a}}rinen.
\newblock Noise-contrastive estimation: {A} new estimation principle for
  unnormalized statistical models.
\newblock In {\em Proceedings of the Thirteenth International Conference on
  Artificial Intelligence and Statistics (AISTATS)}, volume~9, pages 297--304,
  2010.

\bibitem{MnihK13}
Andriy Mnih and Koray Kavukcuoglu.
\newblock Learning word embeddings efficiently with noise-contrastive
  estimation.
\newblock In {\em Advances in Neural Information Processing Systems (NeurIPS)},
  volume~26, pages 2265--2273, 2013.

\bibitem{Sohn16}
Kihyuk Sohn.
\newblock Improved deep metric learning with multi-class n-pair loss objective.
\newblock In {\em Advances in Neural Information Processing Systems (NeurIPS)},
  volume~29, pages 1849--1857, 2016.

\bibitem{SCL}
Prannay Khosla, Piotr Teterwak, Chen Wang, Aaron Sarna, Yonglong Tian, Phillip
  Isola, Aaron Maschinot, Ce~Liu, and Dilip Krishnan.
\newblock Supervised contrastive learning.
\newblock In {\em Advances in Neural Information Processing Systems (NeurIPS)},
  volume~33, pages 18661--18673, 2020.

\bibitem{ZhangTRR21}
Feihu Zhang, Philip H.~S. Torr, Ren{\'{e}} Ranftl, and Stephan~R. Richter.
\newblock Looking beyond single images for contrastive semantic segmentation
  learning.
\newblock In {\em Advances in Neural Information Processing Systems (NeurIPS)},
  pages 3285--3297, 2021.

\bibitem{ZhaoSS2021}
Xiangyun Zhao, Raviteja Vemulapalli, Philip~Andrew Mansfield, Boqing Gong,
  Bradley Green, Lior Shapira, and Ying Wu.
\newblock Contrastive learning for label efficient semantic segmentation.
\newblock In {\em 2021 IEEE/CVF International Conference on Computer Vision
  (ICCV)}, pages 10603--10613, 2021.

\bibitem{ANN}
Piotr Indyk and Rajeev Motwani.
\newblock Approximate nearest neighbors: towards removing the curse of
  dimensionality.
\newblock In {\em Proceedings of the Thirtieth Annual ACM Symposium on Theory
  of Computing}, pages 604--613, 1998.

\bibitem{HPCNN}
Kumar Chellapilla, Sidd Puri, and Patrice Simard.
\newblock High performance convolutional neural networks for document
  processing.
\newblock In {\em Tenth International Workshop on Frontiers in Handwriting
  Recognition}, 2006.

\bibitem{Caffe}
Yangqing Jia, Evan Shelhamer, Jeff Donahue, Sergey Karayev, Jonathan Long, Ross
  Girshick, Sergio Guadarrama, and Trevor Darrell.
\newblock Caffe: Convolutional architecture for fast feature embedding.
\newblock In {\em Proceedings of the ACM International Conference on Multimedia
  (ACM MM)}, pages 675--678, 2014.

\bibitem{cuDNN}
Sharan Chetlur, Cliff Woolley, Philippe Vandermersch, Jonathan Cohen, John
  Tran, Bryan Catanzaro, and Evan Shelhamer.
\newblock {cuDNN}: Efficient primitives for deep learning.
\newblock {\em arXiv preprint arXiv:1410.0759}, 2014.

\bibitem{CCAM}
Jinheng Xie, Jianfeng Xiang, Junliang Chen, Xianxu Hou, Xiaodong Zhao, and
  Linlin Shen.
\newblock C2am: Contrastive learning of class-agnostic activation map for
  weakly supervised object localization and semantic segmentation.
\newblock In {\em Proceedings of the IEEE Conference on Computer Vision and
  Pattern Recognition (CVPR)}, pages 979--988, 2022.

\bibitem{InfoNCE}
Aaron van~den Oord, Yazhe Li, and Oriol Vinyals.
\newblock Representation learning with contrastive predictive coding.
\newblock {\em arXiv preprint arXiv:1807.03748}, 2018.

\bibitem{DAVIS2016}
Federico Perazzi, Jordi Pont-Tuset, Brian McWilliams, Luc Van~Gool, Markus
  Gross, and Alexander Sorkine-Hornung.
\newblock A benchmark dataset and evaluation methodology for video object
  segmentation.
\newblock In {\em Proceedings of the IEEE Conference on Computer Vision and
  Pattern Recognition (CVPR)}, pages 724--732, June 2016.

\bibitem{DAVIS2017}
Jordi Pont-Tuset, Federico Perazzi, Sergi Caelles, Pablo Arbel{\'a}ez, Alex
  Sorkine-Hornung, and Luc Van~Gool.
\newblock The 2017 davis challenge on video object segmentation.
\newblock {\em arXiv preprint arXiv:1704.00675}, 2017.

\bibitem{ResNet}
Kaiming He, Xiangyu Zhang, Shaoqing Ren, and Jian Sun.
\newblock Deep residual learning for image recognition.
\newblock In {\em Proceedings of the IEEE/CVF Conference on Computer Vision and
  Pattern Recognition (CVPR)}, pages 770--778, 2016.

\bibitem{YouTubeVOS}
Ning Xu, Linjie Yang, Yuchen Fan, Dingcheng Yue, Yuchen Liang, Jianchao Yang,
  and Thomas Huang.
\newblock Youtube-vos: A large-scale video object segmentation benchmark.
\newblock {\em arXiv preprint arXiv:1809.03327}, 2018.

\bibitem{PyTorch}
Adam Paszke, Sam Gross, Francisco Massa, Adam Lerer, James Bradbury, Gregory
  Chanan, Trevor Killeen, Zeming Lin, Natalia Gimelshein, Luca Antiga, Alban
  Desmaison, Andreas Kopf, Edward Yang, Zachary DeVito, Martin Raison, Alykhan
  Tejani, Sasank Chilamkurthy, Benoit Steiner, Lu~Fang, Junjie Bai, and Soumith
  Chintala.
\newblock {PyTorch}: {An} imperative style, high-performance deep learning
  library.
\newblock In {\em Advances in Neural Information Processing Systems (NeurIPS)},
  volume~32, pages 8024--8035, 2019.

\bibitem{EdgeFlow}
Yuying Hao, Yi~Liu, Zewu Wu, Lin Han, Yizhou Chen, Guowei Chen, Lutao Chu,
  Shiyu Tang, Zhiliang Yu, Zeyu Chen, and Baohua Lai.
\newblock Edgeflow: Achieving practical interactive segmentation with
  edge-guided flow.
\newblock In {\em Proceedings of the IEEE International Conference on Computer
  Vision Workshops (ICCVW)}, pages 1551--1560, 2021.

\bibitem{EISeg}
Yuying Hao, Yi~Liu, Yizhou Chen, Lin Han, Juncai Peng, Shiyu Tang, Guowei Chen,
  Zewu Wu, Zeyu Chen, and Baohua Lai.
\newblock Eiseg: An efficient interactive segmentation tool based on
  paddlepaddle.
\newblock {\em arXiv preprint arXiv:2210.08788}, 2022.

\bibitem{PerSAM}
Renrui Zhang, Zhengkai Jiang, Ziyu Guo, Shilin Yan, Junting Pan, Xianzheng Ma,
  Hao Dong, Peng Gao, and Hongsheng Li.
\newblock Personalize segment anything model with one shot.
\newblock {\em arXiv preprint arXiv:2305.03048}, 2023.

\bibitem{Matcher}
Yang Liu, Muzhi Zhu, Hengtao Li, Hao Chen, Xinlong Wang, and Chunhua Shen.
\newblock Matcher: Segment anything with one shot using all-purpose feature
  matching.
\newblock In {\em International Conference on Learning Representations (ICLR)},
  2024.

\bibitem{RMem}
Junbao Zhou, Ziqi Pang, and Yu-Xiong Wang.
\newblock Rmem: Restricted memory banks improve video object segmentation.
\newblock In {\em Proceedings of the IEEE Conference on Computer Vision and
  Pattern Recognition (CVPR)}, pages 18602--18611, 2024.

\end{thebibliography}

\end{document}